\documentclass{article}

\usepackage[preprint]{neurips_2026}

\usepackage[utf8]{inputenc} 
\usepackage[T1]{fontenc}    
\usepackage{hyperref}       
\usepackage{url}            
\usepackage{booktabs}       
\usepackage{amsfonts}       
\usepackage{nicefrac}       
\usepackage{microtype}      
\usepackage{xcolor}         

\usepackage{graphicx}
\usepackage{amsmath}
\usepackage{caption}
\usepackage{float}
\usepackage{multirow}
\usepackage{listings}
\usepackage{colortbl}
\usepackage{makecell}
\usepackage{wrapfig}
\usepackage{enumitem}
\usepackage{lipsum} 
\usepackage[ruled,vlined]{algorithm2e}
\usepackage{wrapfig}
\lstdefinestyle{plaintext}{
	basicstyle=\small\ttfamily, 
	breaklines=true,            
	postbreak=\mbox{\textcolor{red}{$\hookrightarrow$}\space}, 
	frame=single,               
	backgroundcolor=\color{gray!5}, 
	numbers=none,               
	tabsize=2,
	aboveskip=1em,
	belowskip=1em
}

\title{DrGait: Biomechanically Grounded Visual Reasoning for Interpretable Clinical Gait Analysis}

\author{%
  Xiangyu Yin, Shiqi Wang, Abrar Alamri, Yasir Aljohani, Weichen Liu, Goeran Fiedler, Wei Gao \\
  University of Pittsburgh \\
  \texttt{\{eric.yin, shw322, aba114, yaa54, weichenliu, gfiedler, weigao\}@pitt.edu} \\
}

\begin{document}

\maketitle

\begin{abstract}
    Current automated gait analysis for clinical applications relies on uninterpretable black-box classifiers. Although Vision-Language Models (VLMs) offer strong reasoning capabilities, applying them directly to gait videos often leads to hallucinations, because they struggle to measure subtle geometric deviations from raw visual contexts. To address this, we introduce \textit{DrGait}, a training-free agentic framework that shifts the VLM's role from a direct visual reasoner to a clinical planner. DrGait decouples semantic reasoning from geometric perception through a structured Triage-Verification-Synthesis (TVS) workflow. Given an input video and a set of basic spatiotemporal metrics, the DrGait agent first performs a heuristic triage to propose diagnostic hypotheses, which are then verified by autonomously calling deterministic biomechanical tools that operate on reconstructed 3D mesh trajectories, segmented 2D pose tracks, and event-centered video evidence. Finally, a closed-loop mechanism recursively updates the agent's reasoning context based on the feedback. By anchoring VLM's reasoning in verifiable geometric and temporal measurements, DrGait reduces hallucinations, achieving competitive diagnostic accuracy while generating transparent and audit-ready clinical reports.
\end{abstract}

 
 \begin{figure}[ht]
 	\setlength{\linewidth}{\textwidth}
 	\setlength{\hsize}{\textwidth}
 	\centering
 	\vspace{-0.15in}
 	\includegraphics[width=0.9 \textwidth]{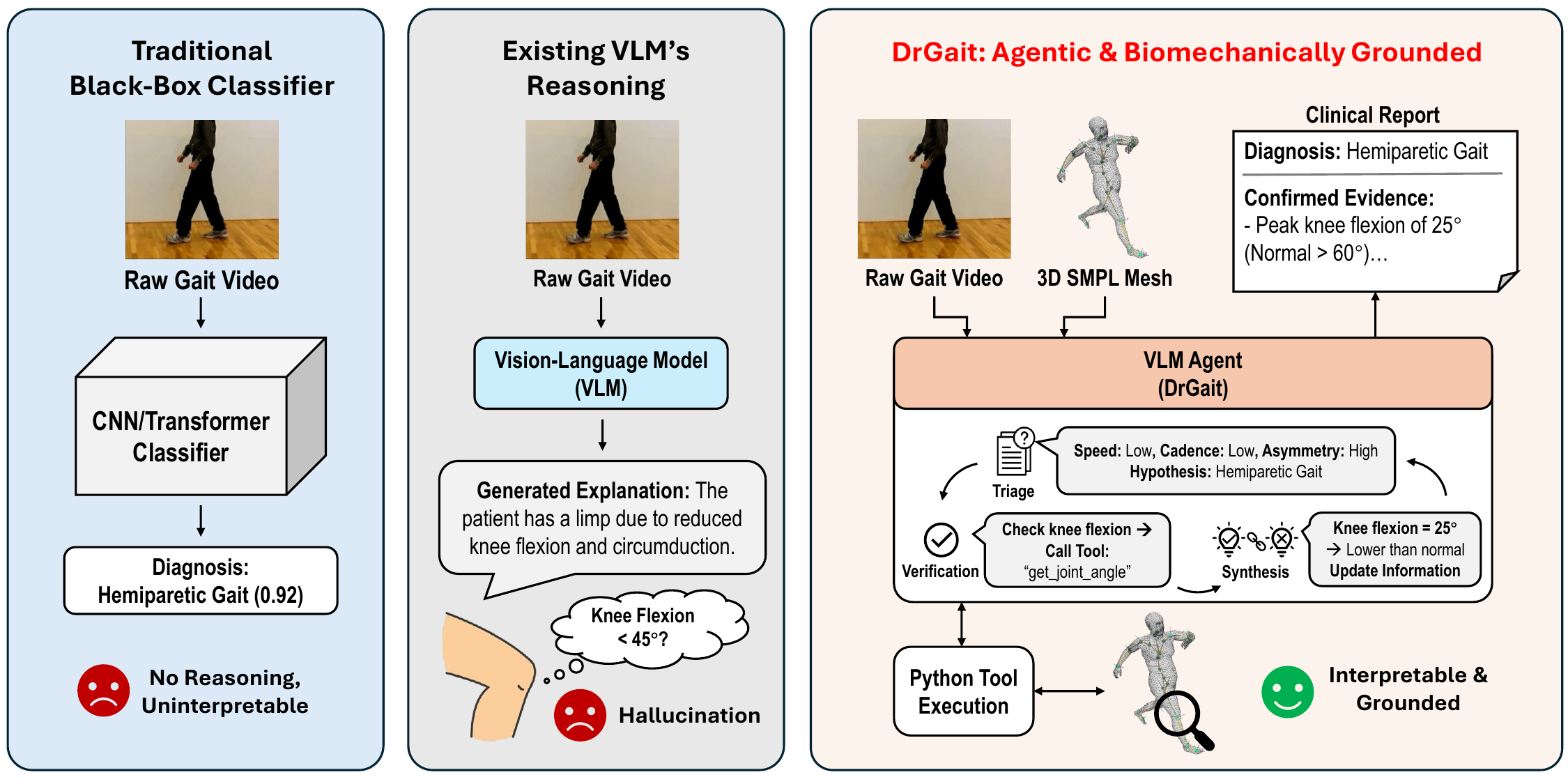} 
 	\vspace{-0.05in}
 	\caption{High-level comparisons between existing methods of gait analysis and DrGait}
 	\vspace{-0.15in}
 	\label{fig:intro}
 \end{figure}
    
\section{Introduction}
\vspace{-0.1in}
\label{sec:intro}

Gait analysis \cite{whittle2014gait, perry2024gait} is crucial in many clinical domains for diagnosis and rehabilitation, 
as deviations from a normal gait pattern can be early pathological indicators of many diseases such as stroke \cite{balaban2014gait} or the Parkinson's disease \cite{mirelman2019gait}. Traditionally, gait analysis relied on visual observations from trained professionals, but is subjective and difficult to standardize \cite{muro2014gait}. While quantitative methods using marker-based motion capture systems exist, their expense and complexity limit their use in many practical scenarios such as home rehabilitation monitoring and fall risk assessment for elderlies \cite{pfister2014comparative}.

Computer vision can be used for automated and objective gait analysis, by extracting quantitative metrics such as skeleton models and Gait Energy Images (GEIs) from gait videos \cite{alharthi2019deep, khan2024deep}. However, existing methods are limited to gait classification (e.g., identifying the ``Parkinsonian gait'') and cannot provide detailed reasoning about the pathological features underlying gait patterns \cite{chavez2022vision, ben2024markerless, trentzsch2021using}. This lack of interpretability hinders their adoption in clinical practice, where understanding the reasons behind diagnosis is critical for disease management and treatment planning \cite{rudin2019stop, tonekaboni2019clinicians}.

Vision-Language Models (VLMs) can provide such interpretability by generating natural language explanations for vision tasks \cite{liu2023visual, moor2023foundation}. However, existing VLMs struggle to measure the subtle geometric deviations from raw visual contexts in gait videos, because their visual encoders are trained with 2D images and lack the capability of correctly perceiving gait characteristics in 3D space. As shown in Figure \ref{fig:intro}, a VLM might correctly identify a patient as ``limping'', but fails to distinguish whether that limp is caused by a deficit in ankle dorsiflexion or an increase in pelvic tilt. Consequently, the VLM reasoning in gait analysis usually contain hallucinations and are hence unreliable.



To address such deficiency in VLM reasoning, existing approaches either fine-tune the VLMs with datasets of specific clinical domains \cite{wang2024enhancing}, or use advanced prompting techniques such as Visual Chain-of-Thought \cite{chen2024visual, hu2024visual, shao2024visual} to explicitly guide the reasoning steps. However, these methods still rely on the VLM's internal visual encoder, which fundamentally lacks the required geometric precision for measuring 3D human motions. Standard visual encoders such as CLIP-based Vision Transformers (ViTs) map 2D image patches into high-level semantic spaces, and compress the continuous spatial coordinates that are necessary for geometric measurements into discrete tokens. As a result, even if being retrained with large-scale 3D human motion data (e.g., Human3.6M \cite{ionescu2013human3} or AMASS \cite{mahmood2019amass}), these text-aligned encoders still suffer from severe numerical hallucinations when calculating the exact kinematic metrics, such as specific joint flexion angles or phase-specific temporal asymmetries.

In this paper, we advocate a fundamental shift: instead of improving the VLM's capability of raw visual perception of geometric measurements, we decouple such visual perception from VLM's reasoning in gait analysis. The VLM is only used for high-level task planning, and numerical measurements from pose, mesh and raw videos are done by specialized computational tools. Based on this idea, we present \textbf{DrGait}, an agentic framework that manages a library of external biomechanical tools, including joint kinematics extractors, gait event detectors, pose-derived symmetry analyzers, etc. The VLM, in turn, decides the task flow of tool calling and uses these numerical gait features for gait analysis reasoning, hence avoiding hallucinations due to visual inaccuracies and ensuring that reasoning is always biomechanically grounded.



DrGait emulates the structured workflow of human physicians through a three-step loop, including Triage, Verification, and Synthesis. 1) \emph{Triage}: DrGait extracts pose or mesh representations from gait videos and calculates basic biomechanical features of gait, such as walking speed and cadence when available. These features are used by the VLM agent to narrow a broad set of pre-defined gait categories into a focused list of candidate hypotheses. 2) \emph{Verification}: The VLM agent validates these hypotheses by calling external tools to extract physical measurements (e.g., joint angles, symmetry index, foot progression, or frontal-plane alignment) from the available geometric representation and event-centered video frames, and the workflow of tool calling is decided by VLM reasoning based on gait characteristics. For example, if the agent suspects a ``foot drop'' issue, it will call a tool to calculate the exact ankle angle at the moment the foot touches ground (i.e., initial contact). 3) \emph{Synthesis}: the VLM agent cross-references the evidence extracted from measurements in Verification with established clinical guidelines, to evaluate the validity of candidate hypotheses and perform a differential diagnosis. If evidence contradicts the initial hypothesis, the agent updates its reasoning context, proposes new hypotheses, and plans a new set of tool calls. This loop continues until evidence supports the final diagnosis without unresolved verification conflicts.

To evaluate DrGait, we conduct extensive experiments on two clinical video datasets: GAVD \cite{ranjan2025computer} for general pathological gaits and ProGait \cite{yin2025progait} for prosthetic gait deviations. Rather than simple forced-choice classification, we task the VLM agent with identifying the correct condition from a comprehensive, open-ended list of clinical categories, and evaluate both the diagnostic accuracy and the semantic correctness of VLM reasoning.
Our results demonstrate that DrGait significantly outperforms VLM baselines, improving the accuracy of gait classification by up to 31\% and approaching the accuracy achieved by human experts. Human evaluations show that DrGait's transparent and tool-backed reports are highly trusted for providing clinical interpretability, being preferred by clinical professionals in 73.2\% of cases on GAVD and 69.6\% on ProGait.

\vspace{-0.1in}
\section{Related Work}
\vspace{-0.1in}
\label{sec:related}
\noindent\textbf{Video-based Gait Analysis:}
Traditional approaches use CNNs and transformers for gait classification \cite{alharthi2019deep, khan2024deep}, but operate as uninterpretable ``black boxes'' \cite{fan2021interpretability}. To improve transparency, recent works connect human motion with LLMs \cite{yang2025bridging, tevet2022motionclip}, but focus on gross action recognition rather than detailed reasoning about pathological deviations. Other efforts fine-tune LLMs to generate rationales for impairment scores \cite{wang2024enhancing, wang2025agir}, but such fine-tuning is expensive and relies on the model's parametric memory, leaving it prone to hallucinations. DrGait avoids fine-tuning entirely, but instead utilizes pre-trained VLMs as active reasoning agents to retrieve verifiable physical data.

\noindent\textbf{VLMs in Medicine:}
Specialized VLMs like Med-PaLM \cite{tu2024towards} and LLaVA-Med \cite{li2023llava} demonstrated impressive capabilities in medical visual question answering (VQA) tasks. However, applying these VLMs directly to gait analysis exposes a critical limitation: standard text-aligned visual encoders (e.g., CLIP \cite{radford2021learning}) are optimized for macroscopic semantic understanding, compressing precise spatial measurements. Consequently, mapping raw videos directly to clinical diagnoses causes VLMs to hallucinate kinematic angles or temporal events. DrGait addresses this flaw by shifting the VLM's role from a direct visual reasoner to a higher-level clinical planner, delegating low-level spatial measurements to precise, deterministic mathematical tools.

\noindent\textbf{Agentic AI and Tool-Use:}
To overcome the limitations of static memory in VLMs, agentic frameworks like Toolformer \cite{schick2023toolformer} and ReAct \cite{yao2022react} empower LLMs to use external tools. While their reasoning structures have evolved from linear Chain-of-Thought \cite{wei2022chain} to complex topologies like Tree-of-Thoughts and Graph-of-Thoughts \cite{besta2024graph}, they remain largely predefined and static. This rigidity is problematic for clinical gait analysis, where diagnostic workflows must dynamically adapt to patient-specific observations (e.g., an asymmetric limp requires a vastly different assessment pathway than a shuffling Parkinsonian gait). DrGait adapts the ReAct paradigm \cite{yao2022react} into a dynamic Triage-Verification-Synthesis (TVS) loop,  and continuously adjusts its reasoning trajectory by calling deterministic biomechanical tools and updating the context based on observations from tools' outputs.

\noindent\textbf{Biomechanical Tools for Gait Analysis:}
Reliable clinical gait diagnosis depends on quantitative biomechanical parameters. Marker-based motion capture can reliably extract high-fidelity human meshes from monocular videos via 3D pose estimation (e.g., WHAM \cite{shin2024wham} and CLIFF \cite{li2022cliff}). Built upon these meshes, deterministic tools such as kinematic event detectors \cite{rueterbories2010methods, hanlon2009real} and ISB-standardized inverse kinematics \cite{wu2002isb} enable the calculation of joint angles and temporal milestones. Existing end-to-end AI methods typically attempt to learn these physical rules from scratch, but DrGait explicitly integrates these deterministic tools as callable functions, whose outputs are then returned to the VLM as auditable biomechanical evidence to bypass the VLM's geometric limitations.

\begin{figure}[t]
	\centering
	\vspace{-0.15in}
	\includegraphics[width=0.88\linewidth]{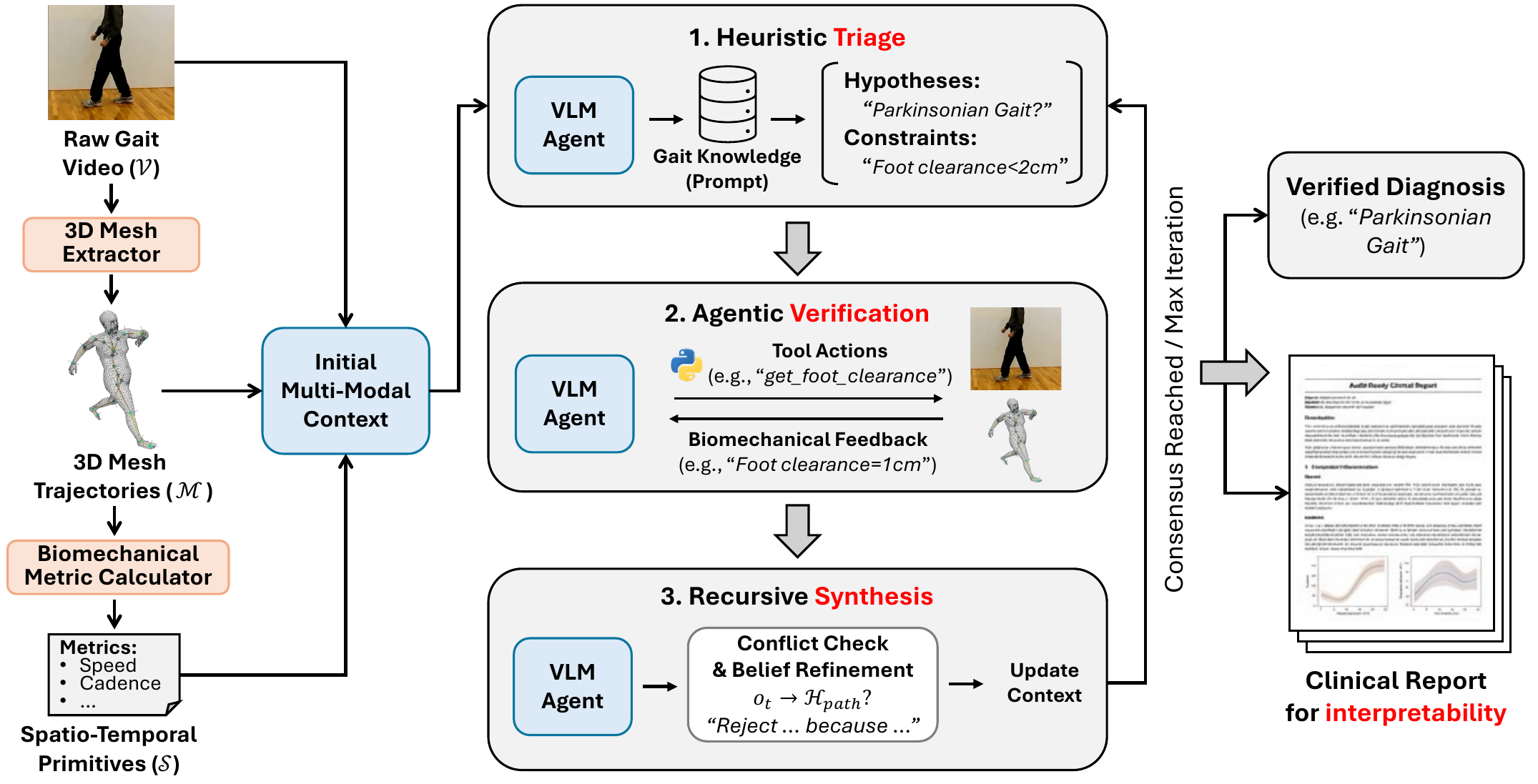}
	\vspace{-0.05in}
	\caption{The overall paradigm of DrGait. To overcome VLMs' weakness of geometric measurements, DrGait replaces direct VLM inference with an iterative Triage-Verification-Synthesis (TVS) workflow. In \textit{Triage}, the VLM agent translates multimodal contexts into candidate hypotheses and formal biomechanical constraints. In \textit{Verification}, the VLM agent orchestrates external biomechanical tools to extract deterministic geometric evidence from mesh trajectories, segmented pose tracks, and event-centered frames. Finally, during \textit{Synthesis}, a closed-loop mechanism recursively refines the agent's belief based on this empirical feedback until a verifiable clinical diagnosis is achieved.}
	\vspace{-0.2in}
	\label{fig:overview}
\end{figure}

\vspace{-0.1in}
\section{Method}
\vspace{-0.1in}
\label{sec:method}

As shown in Figure~\ref{fig:overview}, DrGait is a training-free agentic framework for clinical visual gait analysis, via an interactive and tool-augmented decision-making process. 
The VLM maintains an intermediate working context $c_t$ and repeatedly performs stages of Triage, Verification, and Synthesis (TVS). This TVS reasoning loop serves as the fundamental structure of DrGait, and the differences across various clinical domains and tasks only enter through the input adapter, the injected domain-specific gait knowledge base, and the available tool schemas. In this way, the clinical reasoning framework retains generic by itself, and allows the evidence representation to match tasks in different clinical domains.

\vspace{-0.05in}
\subsection{Multi-Modal Input Context}
\begin{wrapfigure}{r}{3in}
	\centering
	\vspace{-0.25in}
	\includegraphics[width=\linewidth]{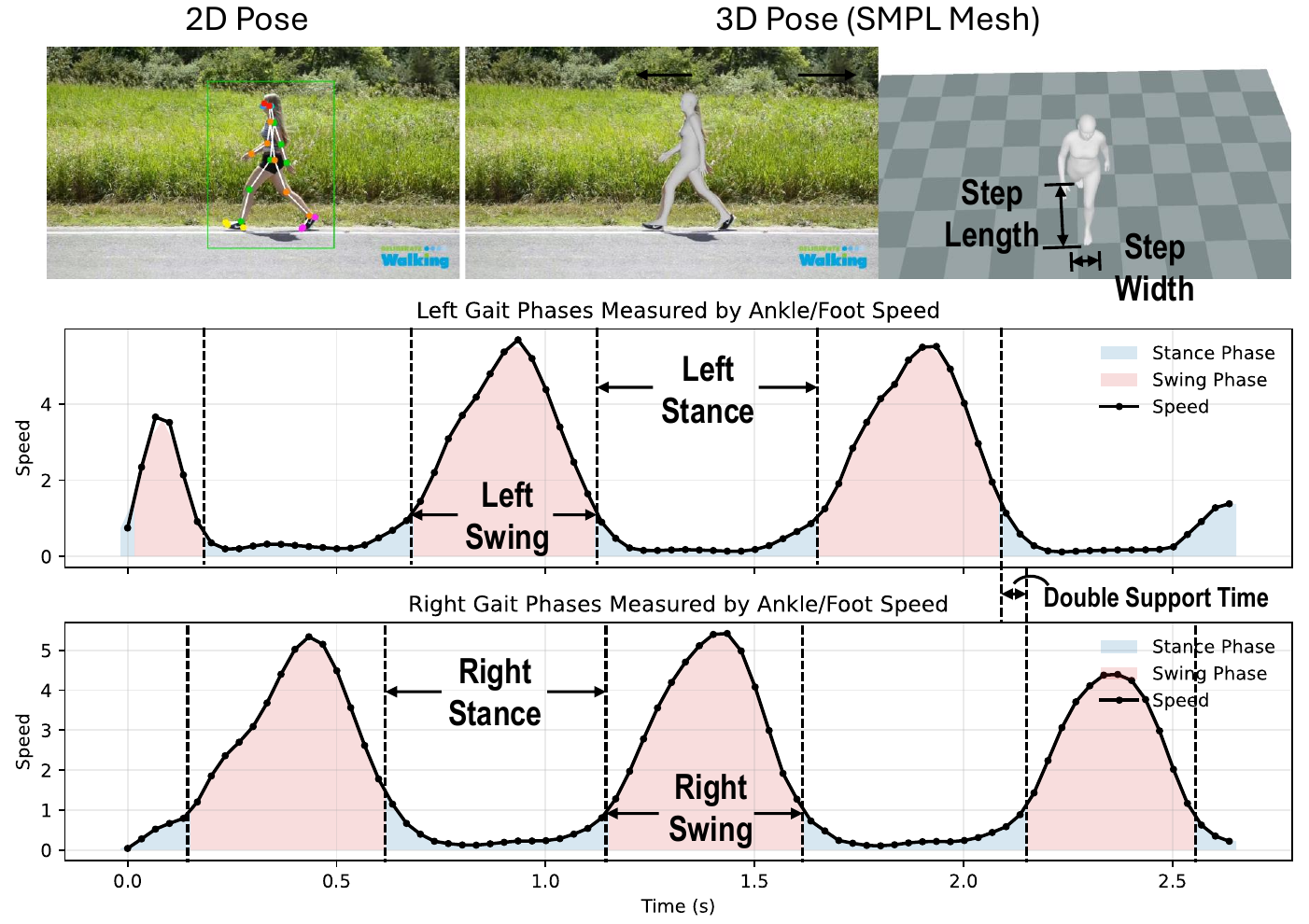}
	\vspace{-0.2in}
	\caption{DrGait's detection of gait phases and extraction of kinematic metrics}
	\vspace{-0.15in}
	\label{fig:gait_phases}
\end{wrapfigure}

\vspace{-0.05in}
\label{sec:input}
Formally, the process by which a naive end-to-end VLM performs gait analysis can be framed as predicting a clinical diagnosis $y \in \mathcal{Y}$ from a raw input video sequence $V \in \mathcal{V}$ via $y = f_\theta(V)$. However, this unconstrained mapping is fundamentally ill-suited for biomechanical assessment, as it promotes implicit decision-making and increases the risk of clinical hallucinations. Instead, DrGait decouples this mapping by initializing an enriched multi-modal context state $c_1$. This starting context provides the VLM with both semantic and geometric scaffolding, which are defined as $c_1 = \{V, \mathcal{P}, \mathcal{S}, \mathcal{D}\}$ and comprise the following:
\vspace{-0.05in}
\begin{itemize}[leftmargin=0.1in]
	\item \textbf{Visual Context ($V$):} The raw video sequence captures high-level cues such as patient effort, posture, assistive devices, prosthetic side, and visible limitations of the camera view.
	\vspace{-0.05in}
	\item \textbf{Pose and Geometry State ($\mathcal{P}$):} The deterministic pose representation available for the given task. When monocular reconstruction is reliable, such pose representation can be reconstructed using the existing tools like WHAM \cite{shin2024wham}, in the form of 3D SMPL mesh and joint trajectory.
	\vspace{-0.05in}
	\item \textbf{Spatiotemporal Primitives ($\mathcal{S}$):} Kinematic metrics extracted from $\mathcal{P}$. For 3D mesh trajectories, we detect gait phases including swing and stance phases, from foot motion in a body-oriented coordinate basis as shown in Figure \ref{fig:gait_phases}, and then computes kinematic metrics including cadence, step length, stride length, step width, and support-time statistics. For segmented 2D trials, metrics are populated with image-plane proxies normalized by body height or shoulder width and aggregated across same-plane segments. These metrics serve as a quantitative ``lab report'' to initialize reasoning.
	\vspace{-0.2in}
	\item \textbf{Gait Knowledge ($\mathcal{D}$):} A task-specific clinical knowledge base injected into the prompt. Each entry defines an exact output label, a correspondingly expected biomechanical pattern, and possible causes. For instance, the knowledge base for one task may define a \textit{Parkinsonian gait} by characteristics like festination, reduced arm swing, stooped posture, and freezing of gait. Another knowledge base for prosthetic gaits could map each fine-grained deviation to a broader general category, such as rotational deviations or knee malalignment. The VLM must choose labels from the knowledge base rather than inventing free-form diagnoses. Details of knowledge bases are in Appendix \ref{sec:gait_dict}.
\end{itemize}

\vspace{-0.1in}
\subsection{Task Formalization via Heuristic Triage}
\vspace{-0.05in}
\label{sec:triage}
In \textit{Triage} stage, the VLM acts as a semantic analyst over the starting context $c_1$. The goal is to identify a set of possible diagnostic hypotheses of gait pattern (e.g., Antalgic, Hemiparetic, Parkinsonian gait, or a prosthetic alignment deviation). To achieve this without hallucination, the agent matches the initial visual and spatiotemporal context against the injected knowledge base $\mathcal{D}$.
Guided by this knowledge, the agent's policy $\pi$ generates a schema-constrained thought $\hat{a}_{task} \in \mathcal{L}$ that decomposes the diagnostic goal into a normality decision, a ranked hypothesis path, and explicit verification objectives. The first substep is a normal-vs-abnormal gate:
\begin{equation}
	g = (z, E_g, U_g) \sim \pi_g(g \mid c_1), \qquad
	z \in \{\text{\ttfamily normal}, \text{\ttfamily abnormal}, \text{\ttfamily uncertain}\},
\end{equation}
where $z$ is the gate classification, $E_g$ contains the visual or metric-grounded evidence used for the gate, and $U_g$ records ambiguity that should be resolved downstream. 
The gate constrains the set of candidate hypothesis. Let $h_{\mathrm{norm}}$ be the normal-gait label, the implementation enforces:
\begin{equation}
	\mathcal{H}_{path} \subseteq
	\begin{cases}
		\{h_{\mathrm{norm}}\}, & z=\text{\ttfamily normal},\\
		\mathcal{H}_{all} \setminus \{h_{\mathrm{norm}}\}, & z=\text{\ttfamily abnormal},\\
		\mathcal{H}_{all}, & z=\text{\ttfamily uncertain}.
	\end{cases}
\end{equation}
In the \texttt{normal} case, $\mathcal{H}_{path}$ must contain exactly one normal hypothesis, but the agent still produces verification objectives to check subtle asymmetry, abnormal step width, small steps, stiff knee, sway, or posture. In the \texttt{abnormal} case, $\mathcal{H}_{path}$ contains only abnormal hypotheses. In the \texttt{uncertain} case, the agent chooses the safest hypothesis path and uses verification objectives to resolve the uncertainty rather than prematurely declaring a diagnosis. The full triage thought is therefore:
\begin{equation}
	\hat{a}_{task} = (s, g, \mathcal{H}_{path}, \mathcal{C}_{bio}) \sim \pi(\hat{a} \mid c_1).
\end{equation}
Here, $\mathcal{H}_{path} \subset \mathcal{H}_{all}$ is a confidence-ranked list of candidate labels from the injected knowledge base, and each element of $\mathcal{C}_{bio}$ specifies a biomechanical objective, the target hypotheses, priority, required observables, and decision rule (e.g., measuring ankle dorsiflexion at initial contact to confirm foot drop, or measuring foot progression angle to distinguish toe-in from circumduction). This thought updates the internal context to $c_2 = (c_1, \hat{a}_{task})$ before any complex computation begins.

\begin{figure}[ht]
	\centering
	\vspace{-0.1in}
	\includegraphics[width=0.85\linewidth]{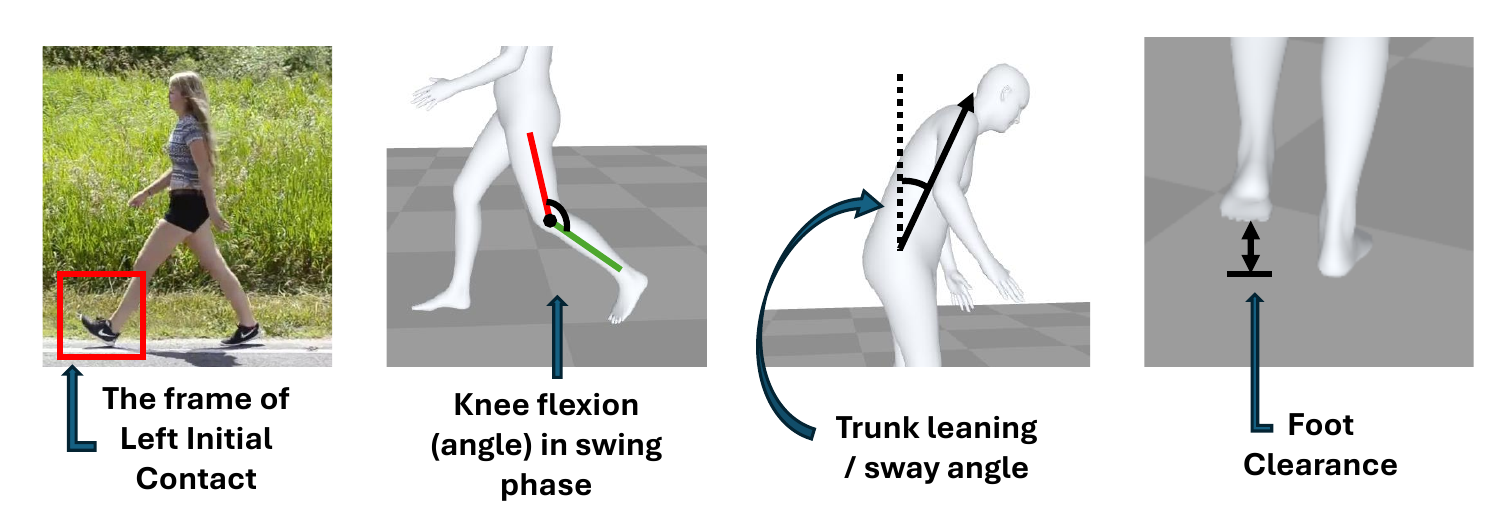}
	\vspace{-0.1in}
	\caption{Examples of external biomechanical tools being used in DrGait}
	\vspace{-0.1in}
	\label{fig:tools}
\end{figure}

\vspace{-0.05in}
\subsection{Agentic Verification and Computation Tools}
\vspace{-0.05in}
\label{sec:verify}
With the task thought injected into the context, the VLM transitions into a constrained planner. Governed by the accumulated context trajectory $c_t$ from the Triage phase, the agent selects task-specific actions $a_t \in \mathcal{A}$. Each action is a schema-constrained function call to an external biomechanical tool:
\begin{equation}
	a_t \sim \pi(a_t \mid c_t).
\end{equation}

To make the verification step easy to audit, we organize the executable tools by the type of clinical evidence they return as follows, and some examples of these tools and the geometric metrics they measure are shown in Figure \ref{fig:tools}. We keep the main text at the evidence-category level and defer function-level details to Appendix \ref{sec:tools}. 
\vspace{-0.05in}
\begin{enumerate}[leftmargin=0.15in]
	\item \textbf{Event evidence.} Captures compact visual evidence at clinically meaningful gait events, such as initial contact, midstance, or toe-off.
	\vspace{-0.05in}
	\item \textbf{Joint motion evidence.} Quantifies phase-specific joint posture and range of motion to test claims such as stiff knee, excessive flexion, or limited ankle motion.
	\vspace{-0.05in}
	\item \textbf{Posture and balance evidence.} Measures trunk, pelvis, arm-swing, and body-center behavior to assess compensatory posture, instability, or reduced upper-body motion.
	\vspace{-0.05in}
	\item \textbf{Foot path and alignment evidence.} Characterizes foot clearance, lateral path deviation, foot orientation, whip direction, and lower-limb alignment for clearance, circumduction, base-width, and prosthetic-alignment hypotheses.
	\vspace{-0.05in}
	\item \textbf{Timing and symmetry evidence.} Summarizes bilateral timing and step asymmetries to evaluate whether one side contributes differently across the gait cycle.
\end{enumerate}
\vspace{-0.05in}

\begin{wrapfigure}{r}{3in}
	\centering
	\vspace{-0.05in}
	\includegraphics[width=0.98\linewidth]{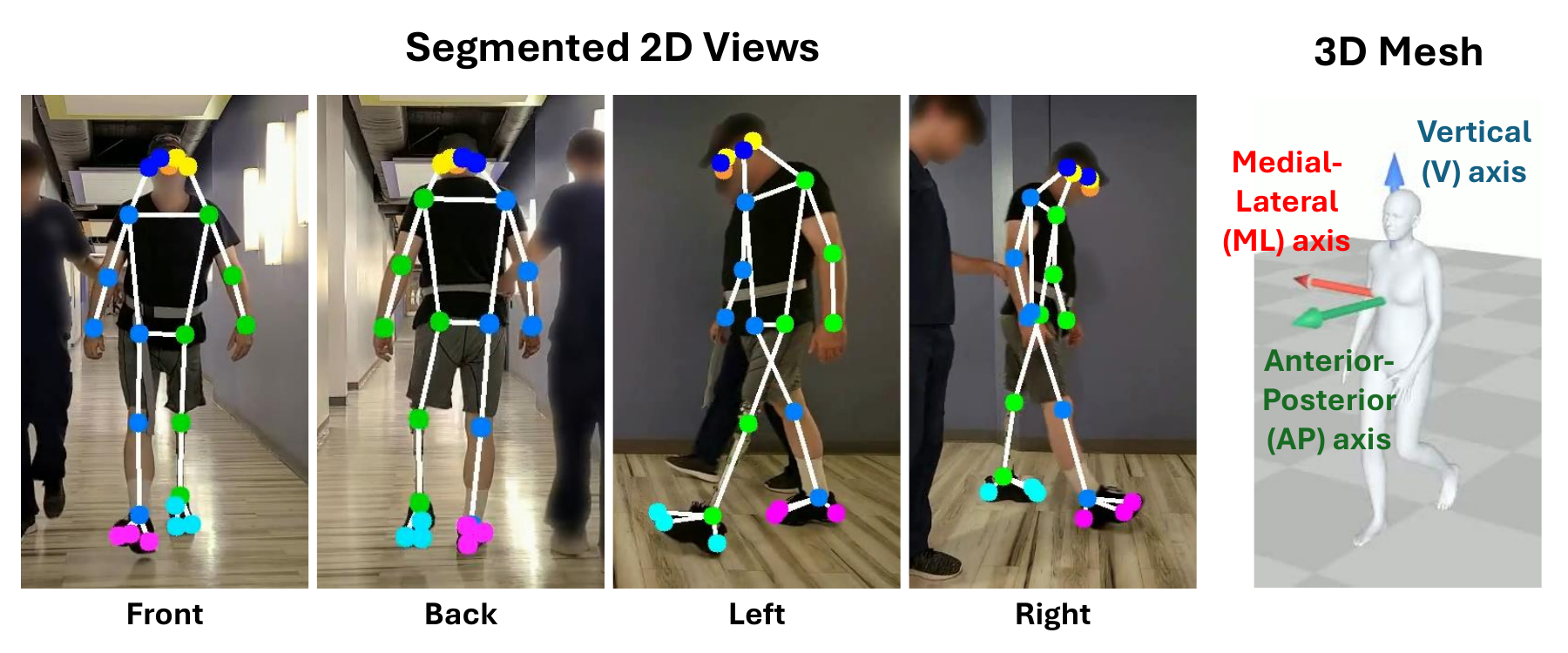}
	\vspace{-0.05in}
	\caption{Visualizations of 2D and 3D contexts}
	\vspace{-0.15in}
	\label{fig:views}
\end{wrapfigure}

Each tool call includes a rationale field (\texttt{why\_used}) explaining which hypothesis or differential question it is meant to resolve. Verification is further restricted to one bounded round of function calling per iteration to prevent uncontrolled tool recursion. Executing $a_t$ interacts with the geometric environment, and returns a deterministic observation feedback $o_t$ containing targeted evidence:
\begin{equation}
	o_t = \text{Execute}(a_t; \mathcal{P}, \mathcal{S}, V)
\end{equation}

This tool-based action step ensures the agent gathers deterministic measurements from external sources rather than hallucinating numerical values. For 3D contexts, anatomical measurements are computed in a stabilized body-oriented basis rather than raw world axes (Figure \ref{fig:views} - right). For segmented 2D contexts, the agent can request a clinically appropriate view: front/back for coronal-plane findings such as base width, pelvic drop, or varus/valgus alignment, and left/right for sagittal-plane findings such as knee flexion, foot clearance, or heel rise (Figure \ref{fig:views} - left). Outputs from 2D pose are explicitly treated as image-plane proxies, not calibrated 3D measurements. The observation is then appended to the state, updating the context to $c_{t+1} = (c_t, a_t, o_t)$.

\vspace{-0.05in}
\subsection{Recursive Synthesis and Belief Refinement}
\vspace{-0.05in}
\label{sec:recursive}
The technical core of DrGait is its closed-loop reasoning mechanism. In the \textit{Synthesis} stage, the agent evaluates the newly retrieved observation $o_t$ against the active clinical hypotheses $\mathcal{H}_{path}$ and the verification objectives $\mathcal{C}_{bio}$. The output is structured to include each objective labeled as supported, refuted, or inconclusive, and each hypothesis is labeled as confirmed, rejected, or inconclusive with a confidence score and explicit supporting/conflicting evidence. The specific fields of the synthesis outputs are as below and the full schema can be found in Appendix \ref{sec:prompts}.

\vspace{-0.1in}
\begin{itemize}[leftmargin=0.1in]
	\item $\mathcal{C}_{bio}$: \textit{"objective"} - biomechanical claim to verify; \textit{"target hypotheses"} - related gait deviation candidates; \textit{"status"} - [supported / refuted / inconclusive]; \textit{"evidence"} - observations from tools.
	\item $\mathcal{H}_{path}$: \textit{"category"} - gait deviation candidates; \textit{"supporting evidence"} - $\mathcal{C}_{bio}$-grounded statements; \textit{"conflicting evidence"} - contradiction; \textit{"status"}: [confirmed / rejected / inconclusive]; \textit{"confidence"}: confidence score of the conclusion.
	\item $SynthesisNotes$: A few sentences summarizing confirmed deviations, rejected distractors, contradictions, and remaining uncertainty.
\end{itemize}
\vspace{-0.1in}

To resolve conflicts or confirm the diagnosis, the agent generates a reasoning trace $\hat{a}_{syn} \in \mathcal{L}$. If the evidence contradicts a hypothesis, this thought acts as a belief refinement (e.g., \textit{``Rejected Hemiparesis because unilateral circumduction is absent; shifting attention to Antalgic residuals''}):
\begin{equation}
	\hat{a}_{syn} \sim \pi(\hat{a} \mid c_{t+1})
\end{equation}

The context is recursively updated with this reasoning trace, such that $c_{t+2} = (c_{t+1}, \hat{a}_{syn})$, allowing dynamica adjustment of future tool calls. This synergistic loop of reasoning ($\hat{a} \in \mathcal{L}$) and acting ($a \in \mathcal{A}$) continues to update $c_t$ until the biomechanical observations and clinical guidelines converge.

The loop terminates only when at least one hypothesis is marked \texttt{confirmed} and no inconclusive verification objective still targets a confirmed hypothesis. If the agent fails to satisfy this criterion after a given maximum number of iterations, DrGait returns the final synthesis produced, with unresolved conflicts or anomalous tool feedback explicitly preserved in the final report.


\vspace{-0.1in}
\section{Experiments}
\vspace{-0.1in}

\subsection{Datasets, Baselines, and Evaluation Protocol}
\vspace{-0.05in}
We evaluate DrGait on two clinical gait datasets. First, we use the GAVD dataset \cite{ranjan2025computer} to evaluate general pathological gait analysis. We include normal gait and five abnormal gait categories: \textit{Antalgic, Myopathic, Cerebral Palsy, Parkinson's}, and \textit{Stroke}. To avoid a simplistic forced-choice setup, the model predicts from an expanded clinical dictionary of 11 gait categories. Multiple fine-grained labels may map back to one dataset class, and distractor categories such as Ataxic Gait are also included. Complete mapping is available in Appendix \ref{sec:gavd_mapping}. After filtering unavailable videos, videos with diagnostic text leakage, severe occlusion/truncation, very short clips, and duplicates, the curated GAVD subset contains 134 gait videos, each of which is processed with RTMW \cite{jiang2024rtmw} for 2D pose and WHAM \cite{shin2024wham} for 3D SMPL reconstruction.

Second, we use the ProGait dataset \cite{yin2025progait} evaluates prosthetic gait analysis using paired frontal and sagittal gait videos. 
The data loader groups paired views into 138 trial-level items and uses annotated straight-walking segments to construct front/back and left/right 2D pose contexts. ProGait evaluation reports general-category accuracy, top-3 general-category accuracy, and fine-grained deviation quality. The mapping between general category and fine-grained deviation is available in Appendix \ref{sec:gait_dict}. Specifically, each predicted gait deviation is mapped to a general prosthetic gait category, and an LLM-as-a-judge scores semantic agreement with the human prosthetic annotation. This judge score rates the semantic similarity between DrGait's generated reports and human annotations from 1 (completely unrelated) to 5 (completely matched).

\begin{table}[t]
	\centering
	\vspace{-0.1in}
	{\fontsize{8}{10}\selectfont
		\begin{tabular}{lcccc}
			\toprule
			\textbf{Method} & \makecell{\textbf{Abnormal (\%)}\\\textbf{(Sensitivity)}} & \makecell{\textbf{Normal (\%)}\\\textbf{(Specificity)}} & \textbf{AUROC} & \makecell{\textbf{Overall}\\\textbf{Accuracy (\%)}} \\
			\midrule
			\textbf{Human Expert} & 99.4 & 100.0 & 0.997 & 99.5 \\
			\midrule
			\rowcolor{gray!25}\vspace{0.05in}\textbf{Training-based} \\
			Pose+PCA+SVM & 82.1 & 18.2 & 0.501 & 69.4 \\
			Pose+LSTM & 85.6 & 27.3 & 0.564 & 73.9 \\
			GaitBase \cite{fan2023opengait}  & 87.3 & 45.5 & 0.784 & 78.4 \\
			GaitGL \cite{lin2022gaitgl} & 86.5 & 36.4 & 0.769 & 76.9 \\
			\midrule
			\rowcolor{gray!25}\vspace{0.05in}\textbf{Training-free} \\
			Gemini-3-Flash & 51.4 & \textbf{100.0} & 0.757 & 61.2 \\
			Gemini-3.1-Pro & 83.2 & 92.6 & 0.879 & 85.1 \\
			GPT-5.4 & 57.9 & 92.6 & 0.752 & 64.9 \\
			Qwen3.6-Plus & 50.5 & 96.0 & 0.733 & 59.1 \\
			\textbf{DrGait (Gemini-3-Flash)} & \textbf{91.6} & 99.3 & \textbf{0.939} & \textbf{92.4} \\
			\bottomrule
	\end{tabular}}
	\vspace{0.05in}
	\caption{Accuracy of binary normality recognition on the GAVD dataset}
	\vspace{-0.25in}
	\label{tab:accuracy_gavd_binary}
\end{table}

For the GAVD dataset, we compare against three baseline families: (1) human expert classification; (2) training-based gait recognition models, including Pose+PCA+SVM, Pose+LSTM, GaitBase \cite{fan2023opengait}, and GaitGL \cite{lin2022gaitgl}; and (3) training-free VLM baselines, including Gemini-3-Flash, Gemini-3.1-Pro, GPT-5.4, and Qwen3.6-Plus. DrGait primarily uses Gemini-3-Flash as the VLM backbone in all experiments. For VLM baselines, we use direct zero-shot prompting with the same clinical reference dictionary. 
Training-based models fall back to a closed six-class GAVD classification task because no training data are available for the distractor classes in the expanded clinical dictionary. Their accuracy therefore cannot be directly compared with the open-dictionary VLM and DrGait settings. For ProGait, we evaluate training-free VLM baselines using the same direct zero-shot prompting protocol with paired video inputs. The complete system and user prompts used for baselines and DrGait are available in Appendix \ref{sec:prompts}.

\begin{table}[ht]	
	\centering
	\vspace{-0.1in}
	{\fontsize{8}{10}\selectfont
		\resizebox{\textwidth}{!}{
			\begin{tabular}{lccccc|c}
				\toprule
				\textbf{Method} & \textbf{Antalgic (\%)} & \makecell{\textbf{Cerebral}\\\textbf{Palsy (\%)}} & \textbf{Myopathic (\%)} & \textbf{Parkinson's (\%)} & \textbf{Stroke (\%)} & \makecell{\textbf{Overall}\\\textbf{Accuracy (\%)}} \\
				\midrule
				\textbf{Human Expert} & 72.7 & 9.7 & 27.6 & 96.1 & 97.4 & 56.1 \\
				\midrule
				\rowcolor{gray!25}\vspace{0.05in}\textbf{Training-based} \\
				Pose+PCA+SVM & 9.1 & 25.0 & 6.9 & 0.0 & 7.7 & 10.3 \\
				Pose+LSTM & 18.2 & 54.2 & 20.7 & 5.9 & 30.8 & 28.0 \\
				GaitBase \cite{fan2023opengait} & 36.4 & 41.7 & 27.6 & 35.3 & 38.5 & 36.4 \\
				GaitGL \cite{lin2022gaitgl} & 27.3 & 41.7 & 34.5 & 23.5 & 30.8 & 32.7 \\
				\midrule
				\rowcolor{gray!25}\vspace{0.05in}\textbf{Training-free} \\
				Gemini-3-Flash & 36.4 & 4.2 & 3.5 & 29.4 & 50.0 & 22.4 \\
				Gemini-3.1-Pro & 9.1 & 16.7 & \textbf{24.1} & 64.7 & 73.1 & 39.3 \\
				GPT-5.4 & 18.2 & 16.7 & 17.2 & 47.1 & 65.4 & 33.6 \\
				Qwen3.6-Plus & 0.0 & 4.2 & 0.0 & 41.2 & 53.9 & 20.6 \\
				\textbf{DrGait (Gemini-3-Flash)} & \textbf{54.6} & \textbf{25.0} & 18.8 & \textbf{94.1} & \textbf{88.5} & \textbf{53.4} \\
				\bottomrule
	\end{tabular}}}
\vspace{0.05in}
	\caption{Accuracy of abnormal gait subtype classification on the GAVD dataset}
	\vspace{-0.25in}
	\label{tab:accuracy_gavd_abnormal}
\end{table}

\vspace{-0.05in}
\subsection{Gait Classification Accuracy}
\vspace{-0.05in}
On the GAVD dataset, we separate normality gait detection from abnormal gait subtype recognition after mapping fine-grained predictions back to the dataset categories. Table~\ref{tab:accuracy_gavd_binary} reports binary normal-vs-abnormal gait recognition performance, and Table~\ref{tab:accuracy_gavd_abnormal} reports accuracy among the five abnormal gait classes. DrGait with Gemini-3-Flash achieves $92.4\%$ accuracy on normality recognition, largely outperforming all the baselines.
Table~\ref{tab:accuracy_gavd_abnormal} shows the similar advantage of DrGait: it reaches $51.4\%$ overall accuracy, improving over Gemini-3-Flash by $29.0\%$ and Gemini-3.1-Pro by $12.1\%$. The largest gains occur in clinically tool-sensitive categories such as Parkinson's ($94.1\%$), Stroke ($88.5\%$), and Antalgic gait ($54.6\%$). 

Compared with human experts, DrGait still trails the clinical upper bound. Human experts achieve near-perfect binary normality recognition ($99.5\%$ overall accuracy) and DrGait reaches $92.4\%$.
For abnormal subtype classification, DrGait reaches $51.4\%$ overall accuracy and approaches the human expert's performance of $56.1\%$, with particularly competitive performance on Parkinson's and Stroke, but weaker performance on heterogeneous categories such as Myopathic and Cerebral Palsy.


\begin{table}[t]
	\centering
	\vspace{-0.05in}
	{\fontsize{8}{10}\selectfont
		\begin{tabular}{lccc}
			\toprule
			\textbf{Method} & \textbf{General Acc. (\%)} & \textbf{Top-3 General Acc. (\%)} & \textbf{LLM-Judge Score} \\
			\midrule
			Gemini-3-Flash & 8.7 & 23.9 & 1.51 \\
			Gemini-3.1-Pro & \textbf{18.1} & 27.5 & 1.68 \\
			GPT-5.4 & 5.1 & 14.5 & 1.44 \\
			Qwen3.6-Plus & 7.3 & 24.6 & 1.62 \\
			\textbf{DrGait (Gemini-3-Flash)} & 16.0 & \textbf{32.6} & \textbf{1.98} \\
			\bottomrule
	\end{tabular}}
	\vspace{0.05in}
	\caption{Gait classification accuracy on the ProGait dataset}
	\vspace{-0.15in}
	\label{tab:accuracy_progait}
\end{table}


On the ProGait dataset, Table~\ref{tab:accuracy_progait} shows that DrGait improves on all reported metrics.
Gemini-3.1-Pro obtains the highest top-1 general-category accuracy ($18.11\%$), but DrGait achieves the best top-3 accuracy and the best semantic agreement with human annotations. This suggests that tool-guided reasoning helps the generated prosthetic-gait report capture relevant deviations even when the exact broad category is not ranked first.

\begin{table}[ht]
	\centering
	{\fontsize{8}{10}\selectfont
		\begin{tabular}{lccccc}
			\toprule
			\textbf{Methods} & \textbf{Diagnostic} & \textbf{Grounding} & \textbf{Usefulness} & \textbf{Overall} & \textbf{Best Win \%} \\
			\midrule
			\rowcolor{gray!25}
			\vspace{0.05in}
			\textbf{GAVD} \\
			Gemini-3-Flash & 2.80 & 2.56 & 2.33 & 2.57 & 8.5 \\
			Gemini-3.1-Pro & 3.23 & 2.82 & 2.60 & 2.88 & 18.3 \\
			\textbf{DrGait (Gemini-3-Flash)} & \textbf{3.68} & \textbf{3.64} & \textbf{3.69} & \textbf{3.67} & \textbf{73.2} \\
			\midrule
			\rowcolor{gray!25}
			\vspace{0.05in}
			\textbf{ProGait} \\
			Gemini-3-Flash & 2.62 & 2.32 & 2.37 & 2.44 & 22.5 \\
			Gemini-3.1-Pro & 2.54 & 2.31 & 2.26 & 2.37 & 27.9 \\
			\textbf{DrGait (Gemini-3-Flash)} & \textbf{2.92} & \textbf{2.97} & \textbf{2.59} & \textbf{2.71} & \textbf{69.6} \\
			\bottomrule
	\end{tabular}}
	\vspace{0.05in}
		\caption{Human evaluations, where scores are average across three humans evaluators.}
		\vspace{-0.1in}
	\label{tab:human_eval}
\end{table}

\subsection{Human Evaluations on Interpretability}
\vspace{-0.05in}
Gait classification accuracy does not capture whether a generated report is clinically useful or interpretable. We then conduct a human expert evaluation of report quality. For each sample, evaluators are shown three reports: one from Gemini-3.1-Pro baseline, one from Gemini-3-Flash baseline, and one from DrGait using Gemini-3-Flash. They rate each report on a 1-to-5 scale across three dimensions: diagnostic correctness, evidence grounding, and clinical usefulness, where 1 indicates worst and 5 indicates best. Evaluators also select the best overall report among the three.

On the GAVE dataset, human evaluation shows that DrGait improves the overall report score to $3.67$, compared with $2.57$ for Gemini-3-Flash and $2.88$ for Gemini-3.1-Pro. 
DrGait is selected as the best overall report in $73.22\%$ of evaluations, compared with $8.47\%$ for Gemini-3-Flash and $18.31\%$ for Gemini-3.1-Pro. On the ProGait dataset, DrGait achieves the highest aggregate overall score ($2.71$) and best-overall selection rate ($49.64\%$).
The ProGait gains are strongest in evidence grounding, reflecting the value of tool-supported measurements for prosthetic-gait reports. Example cases of both datasets and their evaluation forms are in Appendix \ref{sec:gavd_example_cases} and \ref{sec:progait_example_cases}.

\subsection{Efficiency and Cost Analysis}
\vspace{-0.05in}
\label{sec:efficiency}

\begin{wraptable}{r}{3.2in}
	\centering
	\vspace{-0.15in}
	{\fontsize{8}{8.5}\selectfont
	\begin{tabular}{lcc}
		\toprule
		\textbf{Metric} & Gemini-3-Flash & Gemini-3.1-Pro \\
		\midrule
		\textbf{Total End-to-End Latency} & \textbf{11.7s} & \textbf{14.1s}\\
		\midrule
		Input Tokens & 4,762 & 4,762\\
		Output Tokens & 197 & 196\\
		\textbf{Estimated API Cost (USD)} & \textbf{\$0.001} & \textbf{\$0.012}\\
		\bottomrule
	\end{tabular}}
	\vspace{-0.05in}
		\caption{Efficiency and cost per video on GAVD of the VLM baselines}
		\label{tab:efficiency_baseline}
		\vspace{-0.15in}
\end{wraptable}

To evaluate the practical viability of deploying DrGait in real-world clinical settings, we analyze its computational efficiency, latency, and API costs compared to the  zero-shot VLM baselines. This efficiency and cost analysis is based on the GAVD dataset. Results are summarized in Table~\ref{tab:efficiency_baseline} for VLM baselines and Table~\ref{tab:efficiency_drgait} for DrGait.

\textbf{Latency and Agentic Overhead:}
Operating as an autonomous agent naturally introduces a latency trade-off compared to single-shot inference. As shown in the baseline metrics, direct VLM evaluation processes a video in roughly $11.7$ to $14.1$ seconds. In contrast, DrGait requires an average end-to-end latency of $79.9$ seconds (using Gemini-3-Flash) and $107.0$ seconds (using Gemini-3.1-Pro). 

The reasoning duration scales with case complexity. DrGait executes $1.06$ to $1.23$ TVS loops per video, heavily utilizing biomechanical tools ($5.15$ to $5.74$ tool calls per round) to extract reliable kinematic evidence. Despite this overhead, a processing time of under two minutes is highly acceptable for asynchronous clinical diagnosis, especially given the substantial gains in transparency and diagnostic reliability.

\begin{wraptable}{r}{3.2in}
	\centering
	\vspace{-0.1in}
	{\fontsize{8}{9}\selectfont
		\begin{tabular}{lcc}
			\toprule
			\textbf{Metric} & \makecell{\textbf{DrGait}\\Gemini-3-Flash} & \makecell{\textbf{DrGait}\\Gemini-3.1-Pro} \\
			\midrule
			Preprocessing Latency & 50.2s & 50.2s\\
			LLM Agent Latency & 29.7s & 56.8s\\
			\textbf{Total End-to-End Latency} & \textbf{79.9s} & \textbf{107.0s}\\
			\midrule
			Reasoning Rounds  & 1.06 & 1.23\\
			Tool calling & 5.74/round & 5.15/round\\
			Input Tokens & 23,385  & 26,900\\
			Output Tokens & 2,486 & 2,561\\
			\textbf{Estimated API Cost (USD)} & \textbf{\$0.011} & \textbf{\$0.085}\\
			\bottomrule
	\end{tabular}}
	\vspace{-0.05in}
	\caption{Efficiency and cost per video on GAVD of DrGait}
	\label{tab:efficiency_drgait}
\end{wraptable}

\textbf{Token Consumption and Cost-Performance Paradigm:}
Due to the inclusion of detailed clinical knowledge base, the structured action schemas, and the rich tool observation histories, DrGait consumes more tokens than baselines (averaging $\sim23,000 - 27,000$ input tokens versus $\sim4,700$). However, a closer analysis of the API costs reveals a compelling paradigm shift in cost-effectiveness. The direct Gemini-3.1-Pro baseline costs $\$0.012$ per GAVD video and reaches $85.1\%$ binary normality accuracy and $39.3\%$ abnormal-subtype accuracy. DrGait powered by the smaller Gemini-3-Flash model costs slightly less at $\$0.011$ per video while reaching $92.4\%$ binary normality accuracy and $51.4\%$ abnormal-subtype accuracy. This demonstrates a major advantage of DrGait: by offloading the heavy geometric perception to external tools, DrGait empowers a lightweight, low-cost VLM to surpass the raw diagnostic capabilities of a larger direct VLM baseline.

\begin{table}[ht]
	\centering
	\vspace{-0.05in}
	{\fontsize{8}{10}\selectfont
		\begin{tabular}{lcccc}
			\toprule
			\textbf{Ablation} & \makecell{\textbf{GAVD}\\\textbf{Normality Acc. (\%)}} & \makecell{\textbf{GAVD}\\\textbf{Abnormal Acc. (\%)}} & \makecell{\textbf{ProGait}\\\textbf{Top-3 Acc. (\%)}} & \makecell{\textbf{ProGait}\\\textbf{LLM-Judge Score}}\\
			\midrule
			None & 92.4 & 51.4 & 32.6 & 1.78  \\
			Triage-only / no tools & 91.8 & 44.9 & 13.1 & 1.30  \\
			One-pass / no recursion & 91.0 & 43.9 & 34.1 & 1.84   \\
			No normality gate & 85.1 & 33.6 & 35.5 & 1.75  \\
			\bottomrule
	\end{tabular}}
\vspace{0.05in}
\caption{Results of ablation studies}
\vspace{-0.2in}
	\label{tab:ablations}
\end{table}

\subsection{Ablation Studies}
\label{sec:ablation}
\vspace{-0.05in}
We conduct ablation studies on tool-based verification, recursive refinement, and the normality gate. 
Table \ref{tab:ablations} shows that deterministic verification tools are the most important for fine-grained reasoning. Removing tools while keeping the triage stage preserves most of the GAVD normal-vs-abnormal accuracy ($91.8\%$ vs. $92.4\%$), indicating that global gait metrics and visual context are often sufficient for coarse abnormality detection. However, the same ablation reduces GAVD abnormal-subtype accuracy from $51.4\%$ to $44.9\%$, ProGait top-3 accuracy from $32.6\%$ to $13.1\%$, and the ProGait LLM-judge score from $1.78$ to $1.30$. This indicates that verification tools are the main driver of clinically specific differential diagnosis and report grounding.

The one-pass variant, which performs verification without recursive refinement, behaves differently across datasets. On GAVD, recursion improves both normality accuracy ($92.4\%$ vs. $91.0\%$) and abnormal-subtype accuracy ($51.4\%$ vs. $43.9\%$), suggesting that iterative belief updates help resolve ambiguous neurological and musculoskeletal categories. On ProGait, the one-pass variant is slightly higher on top-3 accuracy and LLM-judge score ($34.1\%$ and $1.84$), which suggests that many prosthetic cases are resolved by a single targeted verification round and that additional refinement can sometimes introduce secondary-deviation evidence into the final synthesis.

Finally, removing the normality gate substantially hurts GAVD, reducing normality accuracy to $85.1\%$ and abnormal-subtype accuracy to $33.6\%$. The gate therefore serves an important pruning role for the general pathological-gait setting, where normal and abnormal videos coexist. In contrast, ProGait consists mostly of prosthetic-deviation trials, so removing the gate increases top-3 category retrieval. Overall, tool verification is essential across tasks, recursion is most useful for GAVD-style differential diagnosis, and the normality gate is beneficial when normality itself is part of the decision.

\vspace{-0.1in}
\section{Conclusion}
\vspace{-0.1in}
\label{sec:conclusion}

In this paper, we presented DrGait, a training-free framework for clinically interpretable gait analysis. Through the Triage-Verification-Synthesis loop, a VLM first forms clinically motivated hypotheses, then verifies them with biomechanically-grounded tools, and finally produces an evidence-backed report rather than an unsupported label. DrGait significantly improves VLM's gait analysis capability by pairing general-purpose VLMs with auditable geometric tools.

\setcitestyle{numbers}
\bibliographystyle{unsrt}
\bibliography{main}


\clearpage
\appendix
\appendix

\section{TVS Loop Algorithm}
\label{sec:tvs_algorithm}

\begin{algorithm}[h]
\linespread{0.75}\selectfont
\caption{Current DrGait Triage-Verification-Synthesis (TVS) Loop}
\label{alg:tvs}
\SetAlgoLined
\SetKwFunction{Converged}{Converged}
\SetKwProg{Fn}{Function}{:}{}
\textbf{Input:} visual input $X$ (video(s) or keyframes), pose/geometry context $\mathcal{P}$, metrics $\mathcal{S}$, dictionary $\mathcal{D}$, toolset $\mathcal{A}$, max iterations $N_{max}$ \\
\textbf{Output:} final synthesis $S^*$, final report $R$, audit trace $\mathcal{T}$ \\
$\mathcal{T} \leftarrow [\,]$; $\eta \leftarrow \emptyset$; $\texttt{converged} \leftarrow \textsc{False}$\;
Prepare model visual content from $X$ (upload video views or attach ordered keyframes)\;
\For{$i \leftarrow 1$ \KwTo $N_{max}$}{
    \tcc{Stage 1: schema-constrained triage}
    Build triage context $c_i^{tri} \leftarrow (X, \mathcal{S}, \mathcal{D}, \eta)$\;
    Generate and validate $T_i \sim \pi(\cdot \mid c_i^{tri})$ as \texttt{TriageResponse}\;
    \tcc{$T_i$ contains prosthetic side, normality gate, ranked $H_{path}$, and verification objectives $C_{bio}$}

    \tcc{Stage 2: bounded deterministic verification}
    Ask the agent for function calls $A_i \subseteq \mathcal{A}$ conditioned on $T_i$ and $\eta$\;
    $O_i \leftarrow [\,]$; $I_i \leftarrow [\,]$\;
    \ForEach{$a \in \textsc{FirstK}(A_i, K=10)$}{
        Execute $a$ on $(\mathcal{P}, \mathcal{S}, X)$ using the declared tool schema\;
        Store sanitized tool result, \texttt{why\_used}, arguments, and compact summary in $O_i$\;
        Collect bounded verification images in $I_i$ when tools return event frames\;
    }

    \tcc{Stage 3: differential synthesis}
    Build compact synthesis context $c_i^{syn} \leftarrow (T_i, O_i, I_i, \mathcal{D}, \eta)$\;
    Generate and validate $S_i \sim \pi(\cdot \mid c_i^{syn})$ as \texttt{SynthesisResponse}\;
    \tcc{$S_i$ labels each objective as supported/refuted/inconclusive and each hypothesis as confirmed/rejected/inconclusive}
    Append $(T_i, O_i, S_i)$ and token usage to $\mathcal{T}$\;

    \uIf{\Converged{$S_i$}}{
        $S^* \leftarrow S_i$; $\texttt{converged} \leftarrow \textsc{True}$\;
        \textbf{Break}\;
    }
    \Else{
        $\eta \leftarrow \texttt{synthesis\_notes}(S_i)$\;
        \tcc{Refinement notes are injected into the next triage/verification/synthesis pass}
    }
}
\If{$S^*$ is not set}{
    $S^* \leftarrow$ synthesis output from the last completed iteration\;
}
$R \leftarrow \texttt{final\_report}(S^*)$\;
Write reasoning trace $\mathcal{T}$ and compact final report to disk\;
\textbf{Return} $S^*, R, \mathcal{T}, \texttt{converged}$\;
\BlankLine
\Fn{\Converged{$S$}}{
    $H_c \leftarrow \{h_{\mathrm{cat}} \mid h \in S.H_{\mathrm{path}},\ h_{\mathrm{status}}=\texttt{confirmed}\}$\;
    \If{$H_c = \emptyset$}{\Return \textsc{False}\;}
    \If{any objective in $S.C_{\mathrm{bio}}$ is \texttt{inconclusive} and targets a category in $H_c$}{
        \Return \textsc{False}\;
    }
    \Return \textsc{True}\;
}
\end{algorithm}

\section{GAVD Example Case}
\label{sec:gavd_example_cases}

Below is a complete reasoning trace of DrGait analyzing a GAVD video labeled as Myopathic Gait. This is a representative case in which the initial hypothesis is incorrect, but subsequent verification with biomechanical tools corrects the diagnosis.

DrGait initially favored Parkinsonian gait because the patient showed rapid, short steps and reduced arm swing. Tool verification overturned this interpretation: measured full hip/knee extension and lack of stooped posture refuted Parkinsonian rigidity, while wide base and rhythmic coronal trunk sway supported Waddling/Myopathic gait.

\begin{figure}[h]
	\centering
	\vspace{-0.05in}
	\includegraphics[width=\linewidth]{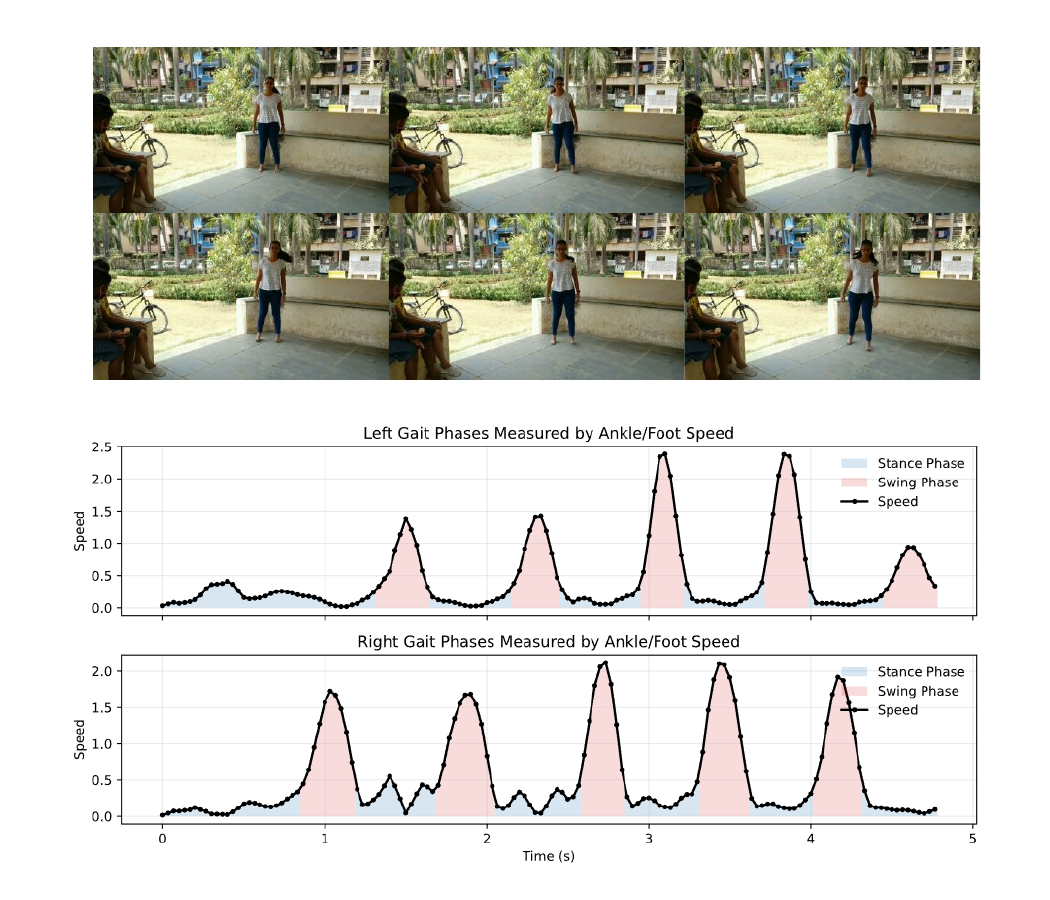}
	\caption{GAVD video example with gait phase detection graph}
	\vspace{-0.05in}
	\label{fig:overview}
\end{figure}

\begin{lstlisting}[style=plaintext, caption={}]
{
  "seq_id": "cljr6cam3003f3n6lzc4mo5ig",
  "video_file_name": "files/gbi6l5tkcq5d",
  "grounding_with_google_search": false,
  "initial_metrics": [
    "mean_cadence_steps_min": 134.0,
    "left_step_length_m": 0.23,
    "right_step_length_m": 0.224,
    "step_length_symmetry_index_pct": 2.63,
    "mean_step_width_m": 0.342,
    "body_sway_amplitude_deg": 9.56,
    "left_stance_phase_pct": 64.3,
    "right_stance_phase_pct": 60.6,
    "stance_time_symmetry_index_pct": 5.95,
    "total_double_support_time_pct": 24.3,
    "step_width_std_m": 0.0219,
    "left_step_length_std_m": 0.0129,
    "right_step_length_std_m": 0.0408,
  ],
  "iterations": [
    {
      "iteration": 1,
      "triage": {
        "response_data": {
          "prosthetic_side": "not_applicable",
          "normality_gate": {
            "classification": "abnormal",
            "evidence": [
              "Severely reduced step lengths (mean < 0.25m) indicating a shuffling pattern.",
              "Increased mean step width of 0.342m, which exceeds the normal threshold of 0.2m.",
              "High cadence of 134 steps/min suggests a rapid, short-stepping (festinating) pattern.",
              "Stance and step length symmetry indices are within normal limits (< 15%), suggesting a bilateral, symmetrical abnormality rather than a unilateral one."
            ],
            "uncertainty": [
              "Visual assessment of postural rigidity and arm swing is limited in the current view.",
              "Need to differentiate between the rhythmic wide-base sway of Waddling gait and the erratic instability of Ataxic gait."
            ]
          },
          "H_path": [
            {
              "id": 1,
              "category": "Parkinsonian (Festinating) Gait",
              "confidence": 0.7,
              "evidence": [
                "Mean step lengths of 0.23m and 0.224m are below the 0.30m threshold for shuffling.",
                "High cadence (134 steps/min) supports a festinating (involuntarily accelerating) pattern.",
                "Symmetrical timing (Stance SI 5.95%) is consistent with bilateral Parkinsonian rigidity."
              ],
              "uncertainty": [
                "Stooped posture and loss of arm swing (upper body rigidity) need visual confirmation or kinematic measurement."
              ]
            },
            {
              "id": 2,
              "category": "Waddling (Myopathic) Gait",
              "confidence": 0.5,
              "evidence": [
                "Increased mean step width (0.342m) is a key feature of the wide base used in myopathic gait.",
                "Symmetrical timing (Stance SI 5.95%) aligns with the bilateral nature of proximal muscle weakness."
              ],
              "uncertainty": [
                "Requires verification of rhythmic lateral trunk sway amplitude to differentiate from Parkinsonian rigidity."
              ]
            },
            {
              "id": 3,
              "category": "Ataxic (Cerebellar) Gait",
              "confidence": 0.3,
              "evidence": [
                "Increased mean step width (0.342m) suggests a compensatory base for instability.",
                "Right step length variability (SD 0.0408m) shows some inconsistency."
              ],
              "uncertainty": [
                "Step width variability (SD 0.0219m) is relatively low, which may contradict the high variability typically seen in Ataxic gait."
              ]
            }
          ],
          "C_bio": [
            {
              "id": 1,
              "objective": "Measure dynamic lateral trunk sway to distinguish rhythmic (Waddling) from rigid (Parkinsonian) or erratic (Ataxic) trunk movement.",
              "target_hypotheses": [
                "Parkinsonian (Festinating) Gait",
                "Waddling (Myopathic) Gait",
                "Ataxic (Cerebellar) Gait"
              ],
              "priority": "high",
              "required_observables": [
                "get_postural_sway_amplitude"
              ],
              "decision_rule": "Large rhythmic sway supports Waddling; minimal/rigid sway supports Parkinsonian; high variability/staggering supports Ataxic."
            },
            {
              "id": 2,
              "objective": "Evaluate trunk and hip flexion angles to identify stooped posture.",
              "target_hypotheses": [
                "Parkinsonian (Festinating) Gait"
              ],
              "priority": "high",
              "required_observables": [
                "trunk_angle",
                "hip_flexion_at_mid_stance"
              ],
              "decision_rule": "Persistent flexion (angles significantly less than 180 degrees) throughout the cycle supports Parkinsonian gait."
            },
            {
              "id": 3,
              "objective": "Verify arm swing amplitude during gait.",
              "target_hypotheses": [
                "Parkinsonian (Festinating) Gait"
              ],
              "priority": "medium",
              "required_observables": [
                "arm_swing_range"
              ],
              "decision_rule": "Bilateral reduction in arm swing supports Parkinsonian rigidity."
            }
          ],
          "ruled_out_distractors": [
            "Hemiparetic (Hemiplegic) Gait",
            "Antalgic Gait",
            "Spastic Diplegic (Scissors) Gait"
          ],
          "triage_notes": "The subject exhibits a symmetrical, wide-based, short-stepping gait with high cadence. This pattern strongly suggests a hypokinetic (Parkinsonian) or proximal weakness (Waddling) profile. Verification will focus on trunk sway characteristics and postural rigidity."
        },
        "token_usage": {
          "input_tokens": 4873,
          "output_tokens": 1165,
          "thought_tokens": 0
        }
      },
      "verification": {
        "tool_trace": [
          {
            "tool": "get_postural_sway_amplitude",
            "why_used": "Measuring dynamic coronal trunk sway to differentiate between rhythmic waddling (high sway) and rigid Parkinsonian (low sway) patterns.",
            "args": {
              "plane": "coronal",
              "segment": "trunk",
              "why_used": "Measuring dynamic coronal trunk sway to differentiate between rhythmic waddling (high sway) and rigid Parkinsonian (low sway) patterns."
            },
            "result_summary": "9.56"
          },
          {
            "tool": "get_postural_sway_amplitude",
            "why_used": "Evaluating sagittal trunk range to assess for rigid, stooped Parkinsonian posture versus dynamic sway.",
            "args": {
              "plane": "sagittal",
              "segment": "trunk",
              "why_used": "Evaluating sagittal trunk range to assess for rigid, stooped Parkinsonian posture versus dynamic sway."
            },
            "result_summary": "4.21"
          },
          {
            "tool": "get_postural_angle",
            "why_used": "Quantifying trunk posture at mid-stance to check for a stooped (flexed) position characteristic of Parkinsonian gait.",
            "args": {
              "segment": "trunk",
              "frame": 12,
              "plane": "sagittal",
              "why_used": "Quantifying trunk posture at mid-stance to check for a stooped (flexed) position characteristic of Parkinsonian gait."
            },
            "result_summary": "{\"angle_deg\": 7.13, \"frame_num\": 12, \"interpretation\": \"Unsigned body-oriented orientation angle. 0 or 180 degrees means the segment is aligned/straight with the reference axis; 90 degrees means perpendicular.\", \"plane\": \"sagittal\", \"pose_space\": \"world_3d\", \"reference_axis\": \"body vertical\", \"segment\": \"trunk\", \"view\": \"front\"}"
          },
          {
            "tool": "get_peak_joint_angle",
            "why_used": "Checking for persistent hip flexion during stance to support a stooped Parkinsonian posture.",
            "args": {
              "phase": "left_stance",
              "joint": "left_hip",
              "why_used": "Checking for persistent hip flexion during stance to support a stooped Parkinsonian posture.",
              "plane": "sagittal"
            },
            "result_summary": "[{\"max_angle\": 180.0, \"min_angle\": 168.37, \"pose_space\": \"world_3d\", \"view\": \"front\"}]"
          },
          {
            "tool": "get_peak_joint_angle",
            "why_used": "Checking for persistent hip flexion during stance to support a stooped Parkinsonian posture.",
            "args": {
              "plane": "sagittal",
              "joint": "right_hip",
              "phase": "right_stance",
              "why_used": "Checking for persistent hip flexion during stance to support a stooped Parkinsonian posture."
            },
            "result_summary": "[{\"max_angle\": 180.0, \"min_angle\": 167.78, \"pose_space\": \"world_3d\", \"view\": \"front\"}]"
          },
          {
            "tool": "get_arm_swing_amplitude",
            "why_used": "Assessing for bilateral reduction in arm swing, a hallmark of Parkinsonian rigidity.",
            "args": {
              "why_used": "Assessing for bilateral reduction in arm swing, a hallmark of Parkinsonian rigidity.",
              "side": "left"
            },
            "result_summary": "{\"max_angle_deg\": -11.29, \"min_angle_deg\": -18.73, \"view\": \"front\"}"
          },
          {
            "tool": "get_arm_swing_amplitude",
            "why_used": "Assessing for bilateral reduction in arm swing, a hallmark of Parkinsonian rigidity.",
            "args": {
              "why_used": "Assessing for bilateral reduction in arm swing, a hallmark of Parkinsonian rigidity.",
              "side": "right"
            },
            "result_summary": "{\"max_angle_deg\": -10.4, \"min_angle_deg\": -15.82, \"view\": \"front\"}"
          },
          {
            "tool": "get_peak_joint_angle",
            "why_used": "Checking for restricted knee flexion to differentiate Parkinsonian rigidity or Stiff-Knee Gait from Neuropathic/Normal flexion.",
            "args": {
              "plane": "sagittal",
              "why_used": "Checking for restricted knee flexion to differentiate Parkinsonian rigidity or Stiff-Knee Gait from Neuropathic/Normal flexion.",
              "joint": "left_knee",
              "phase": "left_swing"
            },
            "result_summary": "[{\"max_angle\": 168.19, \"min_angle\": 156.11, \"pose_space\": \"world_3d\", \"view\": \"front\"}]"
          },
          {
            "tool": "get_joint_angle_at_event",
            "why_used": "Checking for flat-foot strike or foot slap to verify the shuffling pattern associated with Parkinsonian gait.",
            "args": {
              "plane": "sagittal",
              "joint": "left_ankle",
              "why_used": "Checking for flat-foot strike or foot slap to verify the shuffling pattern associated with Parkinsonian gait.",
              "event": "left_initial_contact"
            },
            "result_summary": "[{\"angle\": 109.89, \"frame_num\": 50, \"pose_space\": \"world_3d\", \"step_index\": 0, \"view\": \"front\"}, {\"angle\": 109.83, \"frame_num\": 74, \"pose_space\": \"world_3d\", \"step_index\": 1, \"view\": \"front\"}, {\"angle\": 110.94, \"frame_num\": 97, \"pose_space\": \"world_3d\", \"step_index\": 2, \"view\": \"front\"}]"
          }
        ],
        "token_usage": {
          "input_tokens": 7947,
          "output_tokens": 521,
          "thought_tokens": 0
        }
      },
      "synthesis": {
        "response_data": {
          "C_bio": [
            {
              "id": 1,
              "objective": "Measure dynamic lateral trunk sway to distinguish rhythmic (Waddling) from rigid (Parkinsonian) or erratic (Ataxic) trunk movement.",
              "target_hypotheses": [
                "Parkinsonian (Festinating) Gait",
                "Waddling (Myopathic) Gait",
                "Ataxic (Cerebellar) Gait"
              ],
              "status": "supported",
              "evidence": [
                "get_postural_sway_amplitude (coronal) measured 9.56 deg, which is relatively high for rigid Parkinsonian gait but within rhythmic sway ranges.",
                "get_postural_sway_amplitude (sagittal) measured 4.21 deg, suggesting limited anterior-posterior trunk movement."
              ]
            },
            {
              "id": 2,
              "objective": "Evaluate trunk and hip flexion angles to identify stooped posture.",
              "target_hypotheses": [
                "Parkinsonian (Festinating) Gait"
              ],
              "status": "refuted",
              "evidence": [
                "get_postural_angle (sagittal) found the trunk is only 7.13 degrees from vertical, indicating a largely upright rather than stooped posture.",
                "get_peak_joint_angle for hips showed extension reaching 180 degrees during stance, refuting persistent flexion."
              ]
            },
            {
              "id": 3,
              "objective": "Verify arm swing amplitude during gait.",
              "target_hypotheses": [
                "Parkinsonian (Festinating) Gait"
              ],
              "status": "supported",
              "evidence": [
                "get_arm_swing_amplitude measured very low ranges (~7 deg for left, ~5 deg for right), supporting upper body rigidity."
              ]
            }
          ],
          "H_path": [
            {
              "id": 1,
              "category": "Parkinsonian (Festinating) Gait",
              "supporting_evidence": [
                "Severely reduced step lengths (< 0.25m) and high cadence (134 steps/min).",
                "Significant reduction in bilateral arm swing amplitude (~5-7 deg)."
              ],
              "conflicting_evidence": [
                "Lack of stooped posture (trunk near 7 deg from vertical) and presence of full hip extension (180 deg) in stance."
              ],
              "status": "inconclusive",
              "confidence": 0.4
            },
            {
              "id": 2,
              "category": "Waddling (Myopathic) Gait",
              "supporting_evidence": [
                "Increased mean step width (0.342m).",
                "Presence of rhythmic lateral trunk sway (9.56 deg) and symmetrical timing (SI < 6%)."
              ],
              "conflicting_evidence": [
                "Very short step lengths and high cadence are more typical of Parkinsonian festination than pure myopathic gait."
              ],
              "status": "inconclusive",
              "confidence": 0.4
            },
            {
              "id": 3,
              "category": "Healthy (Normal) Gait",
              "supporting_evidence": [
                "High degree of temporal and spatial symmetry (SI < 10%).",
                "Full hip extension during stance phase."
              ],
              "conflicting_evidence": [
                "Pathologically short step lengths (0.23m) and significantly increased step width (0.342m) relative to height/normative data."
              ],
              "status": "rejected",
              "confidence": 0.1
            }
          ],
          "synthesis_notes": "The gait shows mixed features: Parkinsonian-like shuffling and reduced arm swing, but without the characteristic stooped posture or hip flexion. The wide base and lateral sway overlap with myopathic patterns, but the extreme cadence and short steps suggest a central neurological component. Further clinical assessment of tone and power is required.",
          "final_report": "The analysis identifies an abnormal gait pattern characterized by rapid, very short steps (shuffling) and a widened base of support, though a definitive diagnosis is inconclusive. While the shuffling and reduced arm swing are consistent with a Parkinsonian pattern, the lack of a stooped posture and presence of significant lateral sway also suggest possible proximal weakness or myopathic influence. Recommended next step: Clinical neurological and musculoskeletal evaluation to assess for muscle tone, proximal hip strength, and balance control."
        },
        "token_usage": {
          "input_tokens": 10477,
          "output_tokens": 1036,
          "thought_tokens": 0
        }
      }
    },
    {
      "iteration": 2,
      "triage": {
        "response_data": {
          "prosthetic_side": "not_applicable",
          "normality_gate": {
            "classification": "abnormal",
            "evidence": [
              "Severely reduced step length (mean < 0.25m) suggesting shuffling gait.",
              "Increased mean step width (0.342m) exceeding the normal threshold of 0.2m.",
              "High cadence (134 steps/min) combined with short steps, indicating possible festination.",
              "Visually reduced bilateral arm swing and rigid trunk movement."
            ],
            "uncertainty": [
              "The wide base and lateral sway (9.56 deg) could suggest either a compensatory myopathic pattern or ataxic instability.",
              "While step length is short, step width variability (0.0219m) is relatively low, which may conflict with an ataxic diagnosis."
            ]
          },
          "H_path": [
            {
              "id": 1,
              "category": "Parkinsonian (Festinating) Gait",
              "confidence": 0.65,
              "evidence": [
                "Mean step length is significantly reduced (< 0.30m).",
                "High cadence (134 steps/min) often accompanies the 'festination' (accelerating short steps) of Parkinsonian gait.",
                "Visual observation of reduced bilateral arm swing and rigid 'en bloc' movement."
              ],
              "uncertainty": [
                "Patient lacks the classic stooped posture (trunk flexion) and persists with a wide base, which is atypical for Parkinsonism."
              ]
            },
            {
              "id": 2,
              "category": "Waddling (Myopathic) Gait",
              "confidence": 0.5,
              "evidence": [
                "Increased mean step width (0.342m) to improve lateral stability.",
                "Lateral trunk sway amplitude is measured at 9.56 degrees.",
                "Temporal symmetry is preserved (Stance SI < 15%), consistent with bilateral proximal weakness."
              ],
              "uncertainty": [
                "Myopathic gait typically presents with rhythmic swaying rather than the rapid, shuffling cadence seen here."
              ]
            },
            {
              "id": 3,
              "category": "Ataxic (Cerebellar) Gait",
              "confidence": 0.3,
              "evidence": [
                "Mean step width is > 0.30m (0.342m).",
                "Visual appearance of unsteadiness during initiation and turning."
              ],
              "uncertainty": [
                "Step length and width variability (SD) are relatively low, whereas cerebellar ataxia usually exhibits erratic placement (SD > 0.10m)."
              ]
            }
          ],
          "C_bio": [
            {
              "id": 1,
              "objective": "Verify the presence of a stooped posture or persistent hip/knee flexion.",
              "target_hypotheses": [
                "Parkinsonian (Festinating) Gait",
                "Crouch Gait"
              ],
              "priority": "high",
              "required_observables": [
                "Max hip extension during stance",
                "Max knee extension during stance"
              ],
              "decision_rule": "If max knee angle remains < 160 deg and hips are flexed throughout stance, it supports Parkinsonian or Crouch gait."
            },
            {
              "id": 2,
              "objective": "Measure dynamic trunk sway rhythm and amplitude.",
              "target_hypotheses": [
                "Waddling (Myopathic) Gait",
                "Ataxic (Cerebellar) Gait"
              ],
              "priority": "high",
              "required_observables": [
                "get_postural_sway_amplitude"
              ],
              "decision_rule": "Rhythmic, bilateral sway supports Waddling gait; erratic, unpredictable sway (high variability) supports Ataxic gait."
            },
            {
              "id": 3,
              "objective": "Confirm presence of foot slap or flat-foot strike.",
              "target_hypotheses": [
                "Parkinsonian (Festinating) Gait",
                "Neuropathic (Steppage) Gait"
              ],
              "priority": "medium",
              "required_observables": [
                "Ankle angle at initial contact"
              ],
              "decision_rule": "Ankle angle near neutral or in plantarflexion at initial contact (lack of heel strike) supports Parkinsonian shuffling."
            }
          ],
          "ruled_out_distractors": [
            "Hemiparetic (Hemiplegic) Gait",
            "Antalgic Gait"
          ],
          "triage_notes": "The patient exhibits a shuffling pattern with high cadence and short steps, pointing toward Parkinsonian gait, yet maintains a wide base and lateral sway more typical of myopathic or ataxic conditions. The absence of significant temporal asymmetry rules out Antalgic and Hemiparetic patterns. Primary focus is distinguishing between a hypokinetic central disorder and proximal muscle weakness."
        },
        "token_usage": {
          "input_tokens": 4947,
          "output_tokens": 1165,
          "thought_tokens": 0
        }
      },
      "verification": {
        "tool_trace": [
          {
            "tool": "get_postural_sway_amplitude",
            "why_used": "To distinguish between the rhythmic sway of Waddling gait and the erratic sway of Ataxic gait.",
            "args": {
              "why_used": "To distinguish between the rhythmic sway of Waddling gait and the erratic sway of Ataxic gait.",
              "segment": "trunk",
              "plane": "coronal"
            },
            "result_summary": "9.56"
          },
          {
            "tool": "get_postural_sway_amplitude",
            "why_used": "To check for sagittal instability or stooped posture related to Parkinsonian gait.",
            "args": {
              "why_used": "To check for sagittal instability or stooped posture related to Parkinsonian gait.",
              "segment": "trunk",
              "plane": "sagittal"
            },
            "result_summary": "4.21"
          },
          {
            "tool": "get_peak_joint_angle",
            "why_used": "To determine if the patient maintains persistent knee flexion (Crouch/Parkinsonian) or achieves full extension.",
            "args": {
              "joint": "left_knee",
              "plane": "sagittal",
              "why_used": "To determine if the patient maintains persistent knee flexion (Crouch/Parkinsonian) or achieves full extension.",
              "phase": "left_stance"
            },
            "result_summary": "[{\"max_angle\": 171.56, \"min_angle\": 160.51, \"pose_space\": \"world_3d\", \"view\": \"front\"}]"
          },
          {
            "tool": "get_peak_joint_angle",
            "why_used": "To check for a stooped (flexed) hip posture characteristic of Parkinsonian gait.",
            "args": {
              "why_used": "To check for a stooped (flexed) hip posture characteristic of Parkinsonian gait.",
              "phase": "left_stance",
              "plane": "sagittal",
              "joint": "left_hip"
            },
            "result_summary": "[{\"max_angle\": 180.0, \"min_angle\": 168.37, \"pose_space\": \"world_3d\", \"view\": \"front\"}]"
          },
          {
            "tool": "get_joint_angle_at_event",
            "why_used": "To verify the presence of a flat-foot strike or foot slap, supporting a shuffling/Parkinsonian pattern.",
            "args": {
              "why_used": "To verify the presence of a flat-foot strike or foot slap, supporting a shuffling/Parkinsonian pattern.",
              "event": "left_initial_contact",
              "plane": "sagittal",
              "joint": "left_ankle"
            },
            "result_summary": "[{\"angle\": 109.89, \"frame_num\": 50, \"pose_space\": \"world_3d\", \"step_index\": 0, \"view\": \"front\"}, {\"angle\": 109.83, \"frame_num\": 74, \"pose_space\": \"world_3d\", \"step_index\": 1, \"view\": \"front\"}, {\"angle\": 110.94, \"frame_num\": 97, \"pose_space\": \"world_3d\", \"step_index\": 2, \"view\": \"front\"}]"
          },
          {
            "tool": "get_arm_swing_amplitude",
            "why_used": "To quantify the reduction in arm swing, a key hallmark of Parkinsonian gait.",
            "args": {
              "side": "left",
              "why_used": "To quantify the reduction in arm swing, a key hallmark of Parkinsonian gait."
            },
            "result_summary": "{\"max_angle_deg\": -11.29, \"min_angle_deg\": -18.73, \"view\": \"front\"}"
          },
          {
            "tool": "get_arm_swing_amplitude",
            "why_used": "To check for bilateral reduction in arm swing, supporting Parkinsonian vs unilateral hemiparesis.",
            "args": {
              "side": "right",
              "why_used": "To check for bilateral reduction in arm swing, supporting Parkinsonian vs unilateral hemiparesis."
            },
            "result_summary": "{\"max_angle_deg\": -10.4, \"min_angle_deg\": -15.82, \"view\": \"front\"}"
          },
          {
            "tool": "get_foot_clearance",
            "why_used": "To check for low foot clearance (shuffling) vs high steppage (Neuropathic/Ataxic).",
            "args": {
              "why_used": "To check for low foot clearance (shuffling) vs high steppage (Neuropathic/Ataxic).",
              "phase": "left_swing"
            },
            "result_summary": "0.76"
          }
        ],
        "token_usage": {
          "input_tokens": 8095,
          "output_tokens": 447,
          "thought_tokens": 0
        }
      },
      "synthesis": {
        "response_data": {
          "C_bio": [
            {
              "id": 1,
              "objective": "Verify the presence of a stooped posture or persistent hip/knee flexion.",
              "target_hypotheses": [
                "Parkinsonian (Festinating) Gait",
                "Crouch Gait"
              ],
              "status": "refuted",
              "evidence": [
                "get_peak_joint_angle for the left hip showed a max extension of 180.0 deg, refuting persistent hip flexion.",
                "get_peak_joint_angle for the left knee showed a max extension of 171.56 deg, which is near full extension and refutes the 'persistent flexion' required for Crouch Gait or classic stooped Parkinsonian posture."
              ]
            },
            {
              "id": 2,
              "objective": "Measure dynamic trunk sway rhythm and amplitude.",
              "target_hypotheses": [
                "Waddling (Myopathic) Gait",
                "Ataxic (Cerebellar) Gait"
              ],
              "status": "supported",
              "evidence": [
                "get_postural_sway_amplitude in the coronal plane measured a significant 9.56 deg, supporting the lateral sway characteristic of Waddling gait.",
                "Triage metrics showed a wide step width (0.342m) and low variability (SD 0.0219m), which supports the rhythmic pattern of a Waddling gait over the erratic placement of Ataxic gait."
              ]
            },
            {
              "id": 3,
              "objective": "Confirm presence of foot slap or flat-foot strike.",
              "target_hypotheses": [
                "Parkinsonian (Festinating) Gait",
                "Neuropathic (Steppage) Gait"
              ],
              "status": "supported",
              "evidence": [
                "get_joint_angle_at_event for ankle at initial contact showed angles (~110 deg) indicating a flat-foot strike rather than a heel-first landing.",
                "get_foot_clearance (0.76m) was relatively low, though not extremely restricted, supporting a shuffling-type foot strike."
              ]
            }
          ],
          "H_path": [
            {
              "id": 1,
              "category": "Waddling (Myopathic) Gait",
              "supporting_evidence": [
                "Significant coronal trunk sway (9.56 deg) and wide step width (0.342m) are primary signs.",
                "Temporal symmetry (Stance SI < 15%) is consistent with bilateral proximal weakness."
              ],
              "conflicting_evidence": [
                "High cadence (134 steps/min) and flat-foot strike are more common in hypokinetic gaits than classic myopathy."
              ],
              "status": "confirmed",
              "confidence": 0.7
            },
            {
              "id": 2,
              "category": "Parkinsonian (Festinating) Gait",
              "supporting_evidence": [
                "High cadence (134 steps/min), extremely short steps (< 0.25m), and flat-foot strike.",
                "Bilateral reduction in arm swing (approx 5-7 deg amplitude)."
              ],
              "conflicting_evidence": [
                "Tool measurements confirmed full hip extension (180 deg) and near-full knee extension (171.6 deg), refuting the diagnostic stooped posture.",
                "Wide base (0.342m) is atypical for Parkinsonism."
              ],
              "status": "inconclusive",
              "confidence": 0.4
            },
            {
              "id": 3,
              "category": "Ataxic (Cerebellar) Gait",
              "supporting_evidence": [
                "Increased mean step width (0.342m)."
              ],
              "conflicting_evidence": [
                "Low step-to-step variability in width (SD 0.0219m) refutes the erratic placement required for ataxia."
              ],
              "status": "rejected",
              "confidence": 0.1
            }
          ],
          "synthesis_notes": "The gait exhibits a contradiction between hypokinetic features (high cadence, short steps, flat strike, low arm swing) and mechanical compensations (wide base, significant lateral sway). Tool results refute the stooped posture and sagittal rigidity of Parkinsonian gait, while the rhythmic wide base supports a myopathic or waddling pattern. Clinical assessment should focus on proximal pelvic strength versus bradykinesia.",
          "final_report": "The analysis concludes that the patient demonstrates a Waddling (Myopathic) Gait pattern, characterized by a significantly wide base (0.34m) and pronounced lateral trunk sway (9.56 deg). While some features like a high cadence and flat-foot strike resemble shuffling, the presence of full hip/knee extension and low step variability effectively rules out classic stooped Parkinsonian or Ataxic gaits. This pattern is often associated with proximal muscle weakness around the hips; a clinical assessment of pelvic girdle strength and a referral for physical therapy focusing on core and hip stabilization are recommended."
        },
        "token_usage": {
          "input_tokens": 10211,
          "output_tokens": 1182,
          "thought_tokens": 0
        }
      }
    }
  ],
  "converged": true,
  "num_iterations": 2,
  "token_usage_total": {
    "input_tokens": 46550,
    "output_tokens": 5516,
    "thought_tokens": 0
  },
  "final_synthesis": "The analysis concludes that the patient demonstrates a Waddling (Myopathic) Gait pattern, characterized by a significantly wide base (0.34m) and pronounced lateral trunk sway (9.56 deg). While some features like a high cadence and flat-foot strike resemble shuffling, the presence of full hip/knee extension and low step variability effectively rules out classic stooped Parkinsonian or Ataxic gaits. This pattern is often associated with proximal muscle weakness around the hips; a clinical assessment of pelvic girdle strength and a referral for physical therapy focusing on core and hip stabilization are recommended.",
  "trace_path": "outputs/eval/gavd/default/myopathic/cljr6cam3003f3n6lzc4mo5ig/reasoning_trace.json"
}
\end{lstlisting}

\section{ProGait Example Case}
\label{sec:progait_example_cases}

\begin{figure}[h]
	\centering
	\vspace{-0.05in}
	\includegraphics[width=\linewidth]{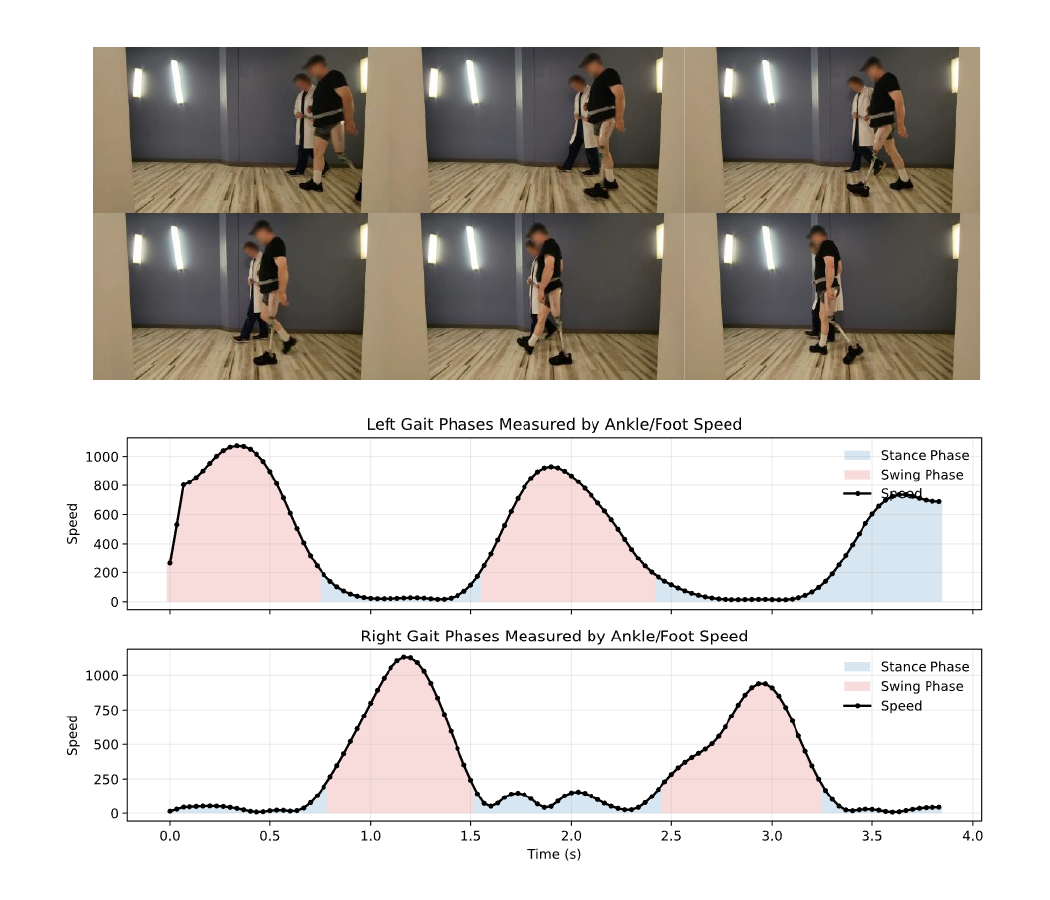}
	\caption{ProGait video example with gait phase detection graph}
	\vspace{-0.05in}
	\label{fig:overview}
\end{figure}

In this ProGait case, DrGait initially interpreted the prosthetic swing pattern as a Lateral Whip. Tool verification corrected the direction of the heel trajectory to a Medial Whip and confirmed a wide base of support through lateral-deviation metrics, while sagittal hip-extension measurements supported associated step-length asymmetry. The final diagnosis shifted to Abducted Gait / Increased Stride Width.

\begin{lstlisting}[style=plaintext]
{
  "seq_id": "outside/2_8_5",
  "video_file_name": {
    "frontal": "files/okqauavo7un3",
    "sagittal": "files/jl60ifv2d12k"
  },
  "grounding_with_google_search": false,
  "initial_metrics": [
    "sagittal_mean_cadence_steps_min": 48.0,
    "sagittal_apparent_step_length_symmetry_index_pct": 11.0,
    "sagittal_stance_time_symmetry_index_pct": 11.5,
    "sagittal_total_double_support_time_pct": 3.7,
    "sagittal_left_knee_angle_min_deg": 145.0,
    "sagittal_left_knee_angle_max_deg": 180.0,
    "sagittal_right_knee_angle_min_deg": 136.0,
    "sagittal_right_knee_angle_max_deg": 180.0,
    "frontal_mean_step_width_shoulder_width_std": 0.0935,
    "frontal_step_width_std_shoulder_width": 0.403,
    "frontal_mean_frontal_knee_angle_deg": 175.0,
  ]
  "iterations": [
    {
      "iteration": 1,
      "triage": {
        "response_data": {
          "prosthetic_side": "right",
          "normality_gate": {
            "classification": "abnormal",
            "evidence": [
              "The user is walking with a manual belt-assist provided by the clinician, indicating significant gait instability.",
              "Visually apparent wide base of support (0.954 mean step width) and trunk shift toward the sound side.",
              "Significant step length asymmetry (11%) and stance time asymmetry (11.5%)."
            ],
            "uncertainty": [
              "Low mean cadence (48 steps/min) may exaggerate or mask certain timing-based deviations."
            ]
          },
          "H_path": [
            {
              "id": 1,
              "category": "Abducted Gait / Increased Stride Width",
              "general_category": {
                "code": 2,
                "name": "Step and Base of Support Deviations"
              },
              "confidence": 0.85,
              "evidence": [
                "Mean step width of 0.954 (nearly equal to shoulder width) is significantly higher than typical prosthetic gait norms.",
                "Frontal view shows the right prosthetic limb carried in a lateral position throughout stance and swing phases."
              ],
              "uncertainty": [
                "Manual belt-assist may be influencing the patient's lateral limb placement or balance strategy."
              ]
            },
            {
              "id": 2,
              "category": "Decreased Prosthetic Stance Time / Short Prosthetic Step",
              "general_category": {
                "code": 3,
                "name": "Step Length and Timing Issues"
              },
              "confidence": 0.75,
              "evidence": [
                "Stance time asymmetry (11.5%) indicates the patient is spending less time on the right prosthetic limb compared to the sound side.",
                "Sagittal view shows a quick, abbreviated stance phase on the right side followed by a rapid step-through of the left limb."
              ],
              "uncertainty": [
                "Asymmetry index of 11.5% is notable but requires stance-phase-specific time measurements to confirm the exact duration on the right limb."
              ]
            },
            {
              "id": 3,
              "category": "Lateral Trunk Lean",
              "general_category": {
                "code": 7,
                "name": "Socket Fit and Stability"
              },
              "confidence": 0.65,
              "evidence": [
                "Visual lean of the torso toward the right prosthetic side during the prosthetic stance phase in the frontal view."
              ],
              "uncertainty": [
                "The clinician's manual gait-belt assist (holding the patient's belt) directly interferes with the natural trunk and pelvic motion."
              ]
            }
          ],
          "C_bio": [
            {
              "id": 1,
              "objective": "Verify the lateral distance of the right foot center from the midline during mid-stance.",
              "target_hypotheses": [
                "Abducted Gait / Increased Stride Width"
              ],
              "priority": "high",
              "required_observables": [
                "Step width measurement",
                "Lateral foot displacement at mid-stance"
              ],
              "decision_rule": "Normalized step width significantly exceeding 0.40 shoulder widths consistently supports Abducted Gait."
            },
            {
              "id": 2,
              "objective": "Compare the duration of the right stance phase vs. left stance phase.",
              "target_hypotheses": [
                "Decreased Prosthetic Stance Time / Short Prosthetic Step"
              ],
              "priority": "high",
              "required_observables": [
                "Right stance time (seconds)",
                "Left stance time (seconds)"
              ],
              "decision_rule": "A stance time for the right side that is at least 10% shorter than the left side confirms the deviation."
            },
            {
              "id": 3,
              "objective": "Measure maximal trunk tilt angle in the frontal plane during right mid-stance.",
              "target_hypotheses": [
                "Lateral Trunk Lean"
              ],
              "priority": "medium",
              "required_observables": [
                "Trunk angle relative to vertical in frontal view"
              ],
              "decision_rule": "A lateral tilt of the trunk exceeding 5 degrees over the prosthetic side supports the deviation, though clinician interference must be noted."
            }
          ],
          "ruled_out_distractors": [
            "Vaulting",
            "Circumduction",
            "Knee Instability (Excessive Knee Flexion)"
          ],
          "triage_notes": "The patient exhibits a highly unstable gait on the right prosthetic side, requiring manual assistance. The primary deviations are a wide base of support (Abducted Gait) and reduced loading time on the prosthesis. Clinician assistance via the gait belt introduces significant uncertainty regarding trunk and pelvic kinematics."
        },
        "token_usage": {
          "input_tokens": 26657,
          "output_tokens": 1146,
          "thought_tokens": 0
        }
      },
      "verification": {
        "tool_trace": [
          {
            "tool": "get_lateral_deviation",
            "why_used": "To measure the lateral displacement of the prosthetic foot during stance to verify Abducted Gait.",
            "args": {
              "view": "front",
              "why_used": "To measure the lateral displacement of the prosthetic foot during stance to verify Abducted Gait.",
              "joint": "right_foot",
              "phase": "right_stance"
            },
            "result_summary": "{\"aggregation\": \"mean_across_segments\", \"mean_max_lateral_deviation_norm\": 0.65, \"per_segment\": {\"front\": {\"interpretation\": \"2D image-plane lateral deviation proxy normalized by shoulder width; not centimeters\", \"max_lateral_deviation_norm\": 0.65, \"normalization\": \"shoulder_width\", \"pose_space\": \"image_2d\", \"scale_px\": 100.0, \"unit\": \"shoulder_width_fraction\", \"view\": \"front\"}}, \"segments\": [\"front\"]}"
          },
          {
            "tool": "calculate_symmetry_index",
            "why_used": "To confirm the magnitude of stance time asymmetry and verify the reduction in loading time on the right prosthetic limb.",
            "args": {
              "why_used": "To confirm the magnitude of stance time asymmetry and verify the reduction in loading time on the right prosthetic limb.",
              "metric_name": "stance_time"
            },
            "result_summary": "{\"aggregation\": \"mean_across_segments\", \"mean_symmetry_index_pct\": 11.39, \"per_segment\": {\"left\": 11.39}, \"segments\": [\"left\"]}"
          },
          {
            "tool": "get_postural_sway_amplitude",
            "why_used": "To quantify the range of trunk motion in the frontal plane to verify Lateral Trunk Lean, despite clinician assistance.",
            "args": {
              "view": "front",
              "plane": "coronal",
              "why_used": "To quantify the range of trunk motion in the frontal plane to verify Lateral Trunk Lean, despite clinician assistance.",
              "segment": "trunk"
            },
            "result_summary": "{\"aggregation\": \"mean_across_segments\", \"mean_amplitude_deg\": 13.9, \"per_segment\": {\"front\": 13.9}, \"segments\": [\"front\"]}"
          },
          {
            "tool": "get_postural_angle",
            "why_used": "To measure the static trunk tilt during right mid-stance for specific evidence of Lateral Trunk Lean.",
            "args": {
              "why_used": "To measure the static trunk tilt during right mid-stance for specific evidence of Lateral Trunk Lean.",
              "segment": "trunk",
              "frame": 10,
              "plane": "coronal",
              "view": "front"
            },
            "result_summary": "{\"aggregation\": \"mean_across_segments\", \"mean_angle_deg\": -2.28, \"per_segment\": {\"front\": {\"angle_deg\": -2.28, \"frame_num\": 10, \"interpretation\": \"Image-plane unsigned orientation angle. 0 or 180 degrees means the segment is aligned/straight with the reference axis; 90 degrees means perpendicular. This is not calibrated 3D posture.\", \"pose_space\": \"image_2d\", \"view\": \"front\"}}, \"segments\": [\"front\"]}"
          },
          {
            "tool": "get_gait_event_frame",
            "why_used": "To visually inspect the lateral position of the limb and trunk alignment at the peak of the prosthetic stance phase.",
            "args": {
              "event": "right_mid_stance",
              "why_used": "To visually inspect the lateral position of the limb and trunk alignment at the peak of the prosthetic stance phase."
            },
            "result_summary": "[{\"image_ref\": \"get_gait_event_frame_image_1\", \"frame_num\": 786, \"step_index\": 0, \"image_source\": \"raw_video\", \"view\": \"front\"}, {\"image_ref\": \"get_gait_event_frame_image_2\", \"frame_num\": 33, \"step_index\": 0, \"image_source\": \"raw_video\", \"view\": \"back\"}, {\"image_ref\": \"get_gait_event_frame_image_3\", \"frame_num\": 60, \"step_index\": 0, \"image_source\": \"raw_video\", \"view\": \"left\"}, {\"image_ref\": \"get_gait_event_frame_image_4\", \"frame_num\": 836, \"step_index\": 1, \"image_source\": \"raw_video\", \"view\": \"front\"}, {\"image_ref\": \"get_gait_event_frame_image_5\", \"frame_num\": 61, \"step_index\": 1, \"image_source\": \"raw_video\", \"view\": \"back\"}]"
          }
        ],
        "token_usage": {
          "input_tokens": 30730,
          "output_tokens": 287,
          "thought_tokens": 0
        }
      },
      "synthesis": {
        "response_data": {
          "C_bio": [
            {
              "id": 1,
              "objective": "Verify the lateral distance of the right foot center from the midline during mid-stance.",
              "target_hypotheses": [
                "Abducted Gait / Increased Stride Width"
              ],
              "status": "supported",
              "evidence": [
                "get_lateral_deviation: Measured a mean max lateral deviation proxy of 0.65 shoulder widths on the right side.",
                "Visual verification (frame 786): Shows the right prosthetic limb significantly lateral to the trunk midline during mid-stance."
              ]
            },
            {
              "id": 2,
              "objective": "Compare the duration of the right stance phase vs. left stance phase.",
              "target_hypotheses": [
                "Decreased Prosthetic Stance Time / Short Prosthetic Step"
              ],
              "status": "supported",
              "evidence": [
                "calculate_symmetry_index: Confirmed a stance time asymmetry of 11.39% favoring the sound side.",
                "Gait metrics: The total stance duration on the right side is notably shorter than the left, consistent with hesitant loading."
              ]
            },
            {
              "id": 3,
              "objective": "Measure maximal trunk tilt angle in the frontal plane during right mid-stance.",
              "target_hypotheses": [
                "Lateral Trunk Lean"
              ],
              "status": "inconclusive",
              "evidence": [
                "get_postural_angle: Found a negligible trunk tilt of -2.28 degrees at frame 10.",
                "get_postural_sway_amplitude: Measured high sway amplitude of 13.9 degrees, but the clinician's manual gait-belt assist prevents reliable isolated assessment of the patient's voluntary trunk lean."
              ]
            }
          ],
          "H_path": [
            {
              "id": 1,
              "category": "Abducted Gait / Increased Stride Width",
              "general_category": {
                "code": 2,
                "name": "Step and Base of Support Deviations"
              },
              "supporting_evidence": [
                "Normalized lateral foot deviation of 0.65 shoulder widths exceeds the triage threshold for abducted gait.",
                "Frontal view confirms the right limb is carried in an abducted position during both swing and stance."
              ],
              "conflicting_evidence": [
                ""
              ],
              "status": "confirmed",
              "confidence": 0.95
            },
            {
              "id": 2,
              "category": "Decreased Prosthetic Stance Time / Short Prosthetic Step",
              "general_category": {
                "code": 3,
                "name": "Step Length and Timing Issues"
              },
              "supporting_evidence": [
                "Verified 11.39% stance time asymmetry indicates the patient spends significantly less time on the right prosthesis.",
                "Visual assessment shows a rapid transition from right stance to sound-side heel strike."
              ],
              "conflicting_evidence": [
                ""
              ],
              "status": "confirmed",
              "confidence": 0.9
            }
          ],
          "synthesis_notes": "Abducted gait and decreased prosthetic stance time are confirmed as the primary deviations. While trunk sway was high, the clinician's manual assist makes it impossible to distinguish between a pathological lean and assisted stabilization. The wide base of support likely serves as a compensatory strategy for the observed instability.",
          "final_report": "The assessment identifies two primary deviations: Abducted Gait / Increased Stride Width and Decreased Prosthetic Stance Time / Short Prosthetic Step. The patient exhibits a significantly wide base of support (0.65 shoulder widths) and spends ~11% less time loading the right prosthesis compared to the sound side, indicating instability. Suggested Action: Review socket fit for medial wall discomfort and check the prosthetic limb length; evaluate the patient's balance confidence and consider socket adduction or linear foot position adjustments to normalize the base of support."
        },
        "token_usage": {
          "input_tokens": 37852,
          "output_tokens": 930,
          "thought_tokens": 0
        }
      }
    }
  ],
  "converged": true,
  "num_iterations": 1,
  "token_usage_total": {
    "input_tokens": 95239,
    "output_tokens": 2363,
    "thought_tokens": 0
  },
  "final_synthesis": "The assessment identifies two primary deviations: Abducted Gait / Increased Stride Width and Decreased Prosthetic Stance Time / Short Prosthetic Step. The patient exhibits a significantly wide base of support (0.65 shoulder widths) and spends ~11% less time loading the right prosthesis compared to the sound side, indicating instability. Suggested Action: Review socket fit for medial wall discomfort and check the prosthetic limb length; evaluate the patient's balance confidence and consider socket adduction or linear foot position adjustments to normalize the base of support.",
  "trace_path": "outputs/eval/progait/progait/outside/2_8_5/reasoning_trace.json"
}
\end{lstlisting}

\section{Gait Knowledge Dictionary}
\label{sec:gait_dict}

\begin{lstlisting}[style=plaintext, caption={GAVD}]
[
  {
    "clinical_category": "Healthy (Normal) Gait",
    "description": "A baseline physiological walking pattern without neuromuscular or musculoskeletal deficits.",
    "key_patterns":[
      "Smooth, rhythmic, and functionally symmetrical walking pattern.",
      "Symmetry Index (SI) < 15% for stance and step length.",
      "NO assistive devices (canes) or medical braces.",
      "Mean step width is typically narrow relative to shoulder width.",
      "Knees flex normally during swing."
    ]
  },
  {
    "clinical_category": "Antalgic Gait",
    "description": "A protective gait adaptation caused by weight-bearing pain in the lower extremity, pelvis, or spine.",
    "key_patterns": [
      "Primary Sign: Severe temporal asymmetry with a notably shortened stance phase on the painful leg to minimize weight-bearing time.",
      "Absence of Hard Neurological Signs: The knee bends normally during swing and there is no circumduction or foot drop.",
      "Step length asymmetry: The step length is typically shortened on the unaffected side (due to rushing to get weight off the painful limb).",
      "Postural compensation: Arm swing and trunk movement may become asymmetrical to aid in balance and unweight the affected side."
    ]
  },
  {
    "clinical_category": "Hemiparetic (Hemiplegic) Gait",
    "description": "An asymmetrical pattern caused by unilateral upper motor neuron damage (e.g., Stroke, CP), often involving compensatory mechanics and assistive devices.",
    "key_patterns": [
      "Asymmetry: Severe spatial and temporal asymmetry is a core hallmark. Do not automatically misattribute this to Antalgic (pain-avoidance) gait.",
      "Assistive Devices: Presence of a unilateral cane/crutch is a strong indicator of hemiparesis, especially if used on the unaffected side to compensate for weakness and instability on the paretic side.",
      "Circumduction: The affected leg swings in a lateral arc.",
      "Compensatory Trunk Sway: Lateral trunk lean is frequently used to assist with foot clearance or balance.",
      "Kinematic Flexibility: While the affected knee often shows reduced swing flexion ('stiff knee'), preserved functional flexion does not rule out hemiparesis if circumduction or cane usage is present.",
      "Upper Limb Synergy: The affected arm is held in a static, flexed posture or demonstrates significantly reduced swing amplitude."
    ]
  },
  {
    "clinical_category": "Parkinsonian (Festinating) Gait",
    "description": "A hypokinetic gait pattern typical of Parkinson's disease, characterized by rigidity, bradykinesia, and postural instability.",
    "key_patterns": [
      "Festination & Shuffling: Severely reduced body-height-normalized step length with a flat-foot strike. The patient takes short, rapid steps that involuntarily accelerate (festination) to catch a forward-leaning center of gravity.",
      "Stooped Posture: The trunk, hips, and knees maintain a persistently flexed, rigid position throughout the entire gait cycle.",
      "Upper Body Rigidity: Bilateral loss of normal arm swing and trunk rotation, causing the patient to move and turn as a single rigid unit ('en bloc' turning).",
      "Initiation Deficits: Prone to 'freezing of gait' (FOG), where the patient experiences a sudden, temporary inability to initiate stepping, especially when starting or turning."
    ]
  },
  {
    "clinical_category": "Waddling (Myopathic) Gait",
    "description": "A bilateral compensated gait pattern caused by proximal pelvic girdle and hip abductor weakness.",
    "key_patterns": [
      "Rhythmic Sway (Bilateral Trendelenburg): Pronounced lateral trunk lean over the stance leg, alternating side-to-side with each step to compensate for pelvic drop.",
      "Wide Base: Increased shoulder-width-normalized step width to improve lateral stability and accommodate the shifting center of mass.",
      "Symmetrical Timing: Because the weakness is bilateral, stance and swing times remain largely symmetrical (Stance SI < 15%).",
      "Exclusion Criteria: Cannot be diagnosed/classified if the patient uses a unilateral cane, as the device artificially alters the bilateral sway dynamics."
    ]
  },
  {
    "clinical_category": "Ataxic (Cerebellar) Gait",
    "description": "An uncoordinated, unsteady gait caused by cerebellar dysfunction, lacking normal motor control.",
    "key_patterns": [
      "Erratic Placement: High normalized step-to-step variability in both step width and length.",
      "Wide Base: Increased shoulder-width-normalized mean step width to compensate for poor balance.",
      "Titubation: Irregular, unpredictable postural sway and staggering, unlike the rhythmic sway of a myopathic gait."
    ]
  },
  {
    "clinical_category": "Neuropathic (Steppage) Gait",
    "description": "A compensatory gait caused by lower motor neuron weakness of the ankle dorsiflexors (foot drop).",
    "key_patterns": [
      "High Steppage: Excessive hip and knee flexion during swing (knee angle drops < 90 deg) to lift the dropping foot clear of the ground.",
      "Foot Slap: Loss of normal heel strike; the foot lands flat or toes-first, often slapping the ground due to lack of eccentric dorsiflexor control.",
      "Orthotic Masking: If the patient wears an AFO, the foot drop and the compensatory high steppage are typically both eliminated."
    ]
  },
  {
    "clinical_category": "Trendelenburg Gait",
    "description": "A mechanically driven gait caused by unilateral weakness of the hip abductors (primarily gluteus medius).",
    "key_patterns": [
      "Uncompensated (Pelvic Drop): The pelvis drops significantly on the swing side when weight is borne on the weak stance leg.",
      "Compensated (Trunk Lean): The patient leans their trunk laterally over the weak stance leg to shift the center of mass and maintain balance.",
      "Differential: Driven entirely by biomechanical weakness, lacking the temporal asymmetry (rushed step) seen in pain-avoidance (Antalgic) gaits."
    ]
  },
  {
    "clinical_category": "Spastic Diplegic (Scissors) Gait",
    "description": "A bilateral upper motor neuron pattern common in Cerebral Palsy, dominated by lower extremity hypertonia.",
    "key_patterns": [
      "Scissoring: Hip adductor spasticity causes a extremely narrow or negative step width, where the knees and feet cross the midline.",
      "Bilateral Stiff Knees: Extensor spasticity restricts normal knee flexion during swing.",
      "Equinus: Calf spasticity forces bilateral toe-walking and lack of heel strike."
    ]
  },
  {
    "clinical_category": "Sensory Ataxic (Stomping) Gait",
    "description": "An uncoordinated gait caused by a loss of proprioception (e.g., peripheral neuropathy, posterior column lesions).",
    "key_patterns": [
      "Stomping Impact: The patient strikes the ground heavily (foot slap/stomp) to maximize residual sensory feedback.",
      "Visual Dependence: The unsteadiness and erratic placement worsen dramatically in the dark or when the patient closes their eyes (positive Romberg sign equivalent).",
      "High Lift: Often accompanied by an exaggerated lifting of the legs due to uncertainty of spatial positioning."
    ]
  },
  {
    "clinical_category": "Crouch Gait",
    "description": "A pathological gait pattern common in spastic Cerebral Palsy, defined by excessive flexion across lower limb joints.",
    "key_patterns": [
      "Persistent Flexion: The hips and knees never fully extend during the stance phase (maximum knee angle remains < 160 deg).",
      "Excessive Ankle Dorsiflexion: Unlike jump gait, crouch gait is typically accompanied by excessive ankle dorsiflexion (calcaneus posture) during stance.",
      "Symmetry: Typically presents bilaterally and symmetrically.",
      "Energy Cost: Highly inefficient, leading to rapid fatigue due to continuous quadriceps activation required to prevent collapsing."
    ]
  }
]
\end{lstlisting}

\begin{lstlisting}[style=plaintext, caption={ProGait}]
[
  {
    "deviation": "Pelvic Drop",
    "general_category": {
      "code": 5,
      "name": "Prosthetic Length Issues"
    },
    "annotation_aliases": [
      "Hip drop",
      "Prosthesis too short",
      "Stepping into a hole"
    ],
    "pattern": "The pelvis on the prosthetic side drops on initial contact as if 'stepping into a hole'.",
    "causes": {
      "patient_related": [
        "Contralateral hip abductor weakness"
      ],
      "prosthetic_related": [
        "Prosthetic limb too short",
        "Residual limb has shrunk relative to the socket",
        "Excessively compliant heel cushion or prosthetic heel keel"
      ]
    }
  },
  {
    "deviation": "Circumduction",
    "general_category": {
      "code": 6,
      "name": "Roll-Over and Clearance Issues"
    },
    "annotation_aliases": [
      "Circumduction gait",
      "Insufficient toe clearance",
      "Lateral arc during swing"
    ],
    "pattern": "The prosthetic limb travels in a lateral arch during the swing phase.",
    "causes": {
      "patient_related": [
        "Inability to adequately flex the knee on the prosthetic side",
        "Abduction contracture",
        "Weak hip flexors",
        "Lack of confidence in flexing the knee",
        "Habit gait pattern or using entire hip/pelvis to initiate gait",
        "Residual limb volume increase"
      ],
      "prosthetic_related": [
        "Prosthesis is physically or functionally too long",
        "Poorly suspended prosthesis",
        "Excessive knee friction",
        "Posterior TKA (Trochanter-Knee-Ankle) alignment"
      ]
    },
    "recommended_actions": [
      "Check socket fit",
      "Shorten prosthesis",
      "Externally rotate the prosthetic knee by adjusting the component inferior to the socket.",
      "Shorten the prosthesis to match the contralateral limb length and level the pelvis.",
      "Shorten the prosthesis to match the sound limb length and level the pelvis.",
      "Adjust prosthetic height"
    ]
  },
  {
    "deviation": "Hip Hike",
    "general_category": {
      "code": 6,
      "name": "Roll-Over and Clearance Issues"
    },
    "annotation_aliases": [
      "Hip hiking",
      "Clearance compensation"
    ],
    "pattern": "The ipsilateral hip rises during the prosthetic swing phase to clear the foot.",
    "causes": {
      "patient_related": [
        "Inability to adequately flex the knee on the prosthetic side",
        "Habit gait pattern",
        "Residual limb volume increase"
      ],
      "prosthetic_related": [
        "Prosthetic limb too long",
        "Poorly suspended prosthesis"
      ]
    }
  },
  {
    "deviation": "Vaulting",
    "general_category": {
      "code": 5,
      "name": "Prosthetic Length Issues"
    },
    "annotation_aliases": [
      "Prosthesis too long",
      "Vaulting gait due to an excessively long prosthesis",
      "Sound-side vaulting",
      "Vaulting gait due to excessive prosthetic length.",
      "Vaulting gait due to an excessively long prosthesis.",
      "Vaulting, secondary to excessive prosthetic length."
    ],
    "pattern": "Ankle plantar flexion and knee extension of the contralateral (sound) limb during the mid-stance phase to assist with prosthetic toe clearance.",
    "causes": {
      "patient_related": [
        "Gait habit or fear of catching the toe",
        "Improper initiation of hip flexors on residual limb",
        "Weak hip flexors on the residual limb"
      ],
      "prosthetic_related": [
        "Prosthetic limb too long (anatomic or functional)",
        "Poor suspension",
        "Excessive knee friction",
        "Posterior TKA alignment"
      ]
    },
    "recommended_actions": [
      "Shorten the prosthesis to match the contralateral limb length and ensure a level pelvis.",
      "Reduce the length of the prosthesis to match the contralateral limb and ensure a level pelvis.",
      "Reduce the prosthetic length to match the contralateral side, thereby leveling the pelvis.",
      "Reduce prosthetic length to match the contralateral limb and achieve a level pelvis."
    ]
  },
  {
    "deviation": "Abducted Gait / Increased Stride Width",
    "general_category": {
      "code": 2,
      "name": "Step and Base of Support Deviations"
    },
    "annotation_aliases": [
      "Abducted gait",
      "Too wide base of support",
      "Excessively wide base of support",
      "Wide step width",
      "Abducted gait resulting in an excessively wide base of support.",
      "Abducted gait, resulting in an excessively wide base of support.",
      "Abducted gait, characterized by an excessively wide base of support.",
      "The patient presents with an abducted gait, resulting in an excessively wide base of support.",
      "Excessively wide base of support, potentially due to excessive socket adduction.",
      "Wide base of support"
    ],
    "pattern": "The prosthetic limb is carried in an abducted position with a widened base of support throughout the swing and stance phase.",
    "causes": {
      "patient_related": [
        "Adaptation for medial compartment knee pain or residual limb pain",
        "Balance impairment, insecurity, or fear of falling",
        "Hip abduction contracture",
        "Weak hip flexors, abductors, or adductors"
      ],
      "prosthetic_related": [
        "Outset prosthetic foot",
        "Prosthesis physically or functionally too long",
        "Medial socket wall too high",
        "Lateral wall insufficient to stabilize the femur (loss of lateral containment)",
        "Excessive knee friction",
        "Excessive plantarflexion"
      ]
    },
    "recommended_actions": [
      "Adjust the socket into adduction by loosening the lateral set screw and tightening the medial set screw inferior to the socket.",
      "Adjust the socket into adduction by loosening the lateral set screw and tightening the medial set screw.",
      "Check socket fit and prosthesis length.",
      "Correct the prosthetic length",
      "Shift the socket laterally",
      "Perform socket abduction by loosening the lateral set screw and tightening the medial set screw."
    ]
  },
  {
    "deviation": "Toe Drag / Reduced Toe Clearance",
    "general_category": {
      "code": 6,
      "name": "Roll-Over and Clearance Issues"
    },
    "annotation_aliases": [
      "Insufficient toe clearance",
      "Toe drag",
      "Reduced toe clearance"
    ],
    "pattern": "The prosthetic toe drags or catches on the ground during the swing phase.",
    "causes": {
      "patient_related": [
        "Muscle weakness of the hip and/or knee flexors",
        "Fatigue",
        "Weak hip extensors",
        "Hip flexion contracture",
        "Toe lever load to unlock the knee does not occur"
      ],
      "prosthetic_related": [
        "Prosthetic limb too long",
        "Excessive plantar flexion of the prosthetic foot",
        "Inadequate suspension or pistoning",
        "Residual limb not fully seated in the socket",
        "Excessive knee friction",
        "Posterior TKA alignment"
      ]
    },
    "recommended_actions": [
      "Reduce swing flexion resistance"
    ]
  },
  {
    "deviation": "Drop Off (Premature Heel Rise)",
    "general_category": {
      "code": 6,
      "name": "Roll-Over and Clearance Issues"
    },
    "annotation_aliases": [
      "Drop off",
      "Premature heel rise",
      "Incomplete roll-over"
    ],
    "pattern": "Premature heel rise at the end of the stance phase, causing a quick 'drop off' onto the contralateral foot.",
    "causes": {
      "patient_related": [
        "Hip flexion contracture",
        "Knee flexion contracture",
        "Knee extensor weakness",
        "Hip extensor weakness"
      ],
      "prosthetic_related": [
        "Foot is too posterior (short toe lever arm)",
        "Foot keel is too soft",
        "Excessive dorsiflexion of the foot",
        "Excessive socket flexion"
      ]
    }
  },
  {
    "deviation": "Foot Slap",
    "general_category": {
      "code": 3,
      "name": "Step Length and Timing Issues"
    },
    "annotation_aliases": [
      "Early foot flat",
      "Foot slap"
    ],
    "pattern": "Rapid, uncontrolled plantarflexion of the foot immediately after initial heel contact.",
    "causes": {
      "patient_related": [
        "Poor eccentric dorsiflexion control (if observed on the intact lower limb)"
      ],
      "prosthetic_related": [
        "Posterior (plantar flexion) bumper is too soft",
        "Insufficient dampening of plantar flexion in a hydraulic ankle"
      ]
    }
  },
  {
    "deviation": "Inadequate Heel Off",
    "general_category": {
      "code": 6,
      "name": "Roll-Over and Clearance Issues"
    },
    "annotation_aliases": [
      "Incomplete roll-over",
      "Insufficient heel off",
      "Delayed roll-over"
    ],
    "pattern": "Insufficient heel rise at the end of the stance phase.",
    "causes": {
      "patient_related": [
        "Reduced confidence in putting weight through the toe of the prosthesis"
      ],
      "prosthetic_related": [
        "Foot is too anterior (long toe lever arm)",
        "Foot keel is too stiff"
      ]
    }
  },
  {
    "deviation": "Decreased Prosthetic Stance Time / Short Prosthetic Step",
    "general_category": {
      "code": 3,
      "name": "Step Length and Timing Issues"
    },
    "annotation_aliases": [
      "Short prosthetic step",
      "Reduced prosthetic stance time"
    ],
    "pattern": "The total duration of the stance phase on the prosthetic limb is reduced, resulting in a short, quick step with the contralateral foot.",
    "causes": {
      "patient_related": [
        "Residual limb pain or discomfort",
        "Patient insecurity or poor balance",
        "Weak hip muscles",
        "Musculoskeletal pain in proximal structures"
      ],
      "prosthetic_related": [
        "Poorly fitting prosthetic socket",
        "Prosthetic foot alignment issues (e.g., toe lever too short)",
        "Unstable knee",
        "Excess socket flexion or ankle dorsiflexion",
        "Poor suspension"
      ]
    }
  },
  {
    "deviation": "Uneven Timing / Long Prosthetic Step",
    "general_category": {
      "code": 3,
      "name": "Step Length and Timing Issues"
    },
    "annotation_aliases": [
      "Asymmetric step length",
      "Uneven step length",
      "Long prosthetic step",
      "Uneven timing"
    ],
    "pattern": "Hesitation in gait timing from prosthetic mid-stance to sound side heel strike, causing the sound limb to take an unusually long step.",
    "causes": {
      "patient_related": [
        "Hip flexion contracture",
        "Pain on the sound limb side"
      ],
      "prosthetic_related": [
        "Excessive ankle plantarflexion",
        "Insufficient socket flexion",
        "Long or excessively stiff toe lever"
      ]
    }
  },
  {
    "deviation": "Knee Instability (Excessive Knee Flexion)",
    "general_category": {
      "code": 4,
      "name": "Knee Instability and Malalignment"
    },
    "annotation_aliases": [
      "Knee instability",
      "Excessive knee flexion",
      "Buckling"
    ],
    "pattern": "Excessive knee flexion on the prosthetic side in early stance; the knee tends to buckle, and the patient feels like they are 'walking downhill'.",
    "causes": {
      "patient_related": [
        "Knee flexion contracture",
        "Quadriceps weakness",
        "Weak hip extensors",
        "Short residual limb causing excessive play"
      ],
      "prosthetic_related": [
        "Excessive foot dorsiflexion",
        "Excessive socket flexion (forward tilt on pylon)",
        "Posterior translation of the foot/pylon (heel lever too long)",
        "Excessively hard heel cushion or prosthetic heel",
        "Knee axis of rotation is too anterior"
      ]
    }
  },
  {
    "deviation": "Knee Too Extended",
    "general_category": {
      "code": 4,
      "name": "Knee Instability and Malalignment"
    },
    "annotation_aliases": [
      "Hyperextended knee",
      "Knee hyperextension",
      "Knee too extended"
    ],
    "pattern": "Excessive knee extension during the stance phase; the patient feels like they are 'walking uphill'.",
    "causes": {
      "patient_related": [
        "Inadequate knee flexion range of motion or knee extension contracture",
        "Compensation for quadriceps weakness",
        "Compensation for hip flexion contracture"
      ],
      "prosthetic_related": [
        "Excessively compliant prosthetic heel cushion or forefoot keel too rigid",
        "Inadequate socket flexion",
        "Excessive plantar flexion of the prosthetic foot",
        "Anterior translation of the prosthetic foot/pylon (excess toe lever)"
      ]
    },
    "recommended_actions": [
      "Flex the socket",
      "Decrease ankle plantarflexion"
    ]
  },
  {
    "deviation": "Excessive Valgus Moment at the Knee",
    "general_category": {
      "code": 4,
      "name": "Knee Instability and Malalignment"
    },
    "annotation_aliases": [
      "Excessive valgus",
      "Knee valgus asymmetry",
      "Medial thrust",
      "Excessive knee valgus, potentially causing a medial whip at higher gait speeds."
    ],
    "pattern": "A medial thrust of the proximal socket is observed during the mid-stance phase.",
    "causes": {
      "patient_related": [
        "Short residual limb causing poor stabilization of the prosthetic socket",
        "Ligamentous instability"
      ],
      "prosthetic_related": [
        "Excessive lateral translation of the prosthetic pylon/foot",
        "Excessive valgus angulation at the prosthetic socket pylon junction",
        "Socket too loose"
      ]
    },
    "recommended_actions": [
      "Reduce the knee valgus angle."
    ]
  },
  {
    "deviation": "Excessive Varus Moment / Lateral Shift",
    "general_category": {
      "code": 4,
      "name": "Knee Instability and Malalignment"
    },
    "annotation_aliases": [
      "Knee varus asymmetry",
      "Knee varus",
      "Varus deformity",
      "Lateral thrust"
    ],
    "pattern": "A lateral thrust of the proximal socket occurs in mid-stance phase.",
    "causes": {
      "patient_related": [
        "Inadequate balance",
        "Weak hip abductors",
        "Narrow gait base"
      ],
      "prosthetic_related": [
        "Excessive inset of the prosthetic foot",
        "Excessive abduction of the socket (brim tilted medially)",
        "Socket too loose",
        "Loss of frontal plane containment",
        "Short prosthesis"
      ]
    }
  },
  {
    "deviation": "Lateral Trunk Lean",
    "general_category": {
      "code": 7,
      "name": "Socket Fit and Stability"
    },
    "annotation_aliases": [
      "Lateral instability",
      "Compensated Trendelenburg",
      "Trunk lean"
    ],
    "pattern": "The trunk bends laterally over the prosthesis (compensated Trendelenburg) during the prosthetic stance phase.",
    "causes": {
      "patient_related": [
        "Weak hip abductors",
        "Pain at the distal lateral residual limb",
        "Gait habit or inadequate balance",
        "Short residual limb",
        "Abduction contracture"
      ],
      "prosthetic_related": [
        "Inadequate adduction of the socket",
        "Prosthesis too short",
        "Outset foot",
        "Medial wall too high causing pain",
        "Ineffective lateral socket containment"
      ]
    }
  },
  {
    "deviation": "Trunk Lordosis",
    "general_category": {
      "code": 7,
      "name": "Socket Fit and Stability"
    },
    "annotation_aliases": [
      "Socket fit issue",
      "Lumbar lordosis"
    ],
    "pattern": "Increased lumbar arch (hyperlordosis) observed during the stance phase.",
    "causes": {
      "patient_related": [
        "Compensation for tight hip flexors",
        "Weak hip extensors",
        "Weak abdominal/core muscles",
        "Gait habit"
      ],
      "prosthetic_related": [
        "Insufficient socket flexion",
        "Posterior wall promotes anterior pelvic tilt"
      ]
    }
  },
  {
    "deviation": "Forward Trunk Lean",
    "general_category": {
      "code": 7,
      "name": "Socket Fit and Stability"
    },
    "annotation_aliases": [
      "Forward trunk lean",
      "Stability compensation"
    ],
    "pattern": "The trunk leans forward significantly during the stance phase.",
    "causes": {
      "patient_related": [
        "Hip flexion contracture",
        "Lumbar pain aggravated by lumbar extension",
        "Compensation to prevent knee instability"
      ],
      "prosthetic_related": [
        "Knee is in an unstable alignment",
        "Stance control mechanism of the knee is insufficient"
      ]
    }
  },
  {
    "deviation": "External Rotation of Foot on Initial Contact",
    "general_category": {
      "code": 1,
      "name": "Rotational Deviations"
    },
    "annotation_aliases": [
      "Toe-out asymmetry",
      "Externally rotated foot",
      "External foot rotation"
    ],
    "pattern": "The prosthetic foot externally rotates immediately upon heel strike.",
    "causes": {
      "patient_related": [
        "Poor residual limb muscle control (decreased stability)",
        "Prosthesis improperly donned in external rotation"
      ],
      "prosthetic_related": [
        "Heel cushion or plantar-flexion bumper is too stiff",
        "Too much toe-out alignment built into the prosthesis"
      ]
    }
  },
  {
    "deviation": "Medial Whip",
    "general_category": {
      "code": 1,
      "name": "Rotational Deviations"
    },
    "annotation_aliases": [
      "Medial whip",
      "Externally rotated knee",
      "Externally rotated prosthetic knee"
    ],
    "pattern": "Abrupt medial movement of the prosthetic heel during the swing phase.",
    "causes": {
      "patient_related": [
        "Socket donned in external rotation",
        "Gait habit (external rotation of the hip at toe-off)",
        "Insufficient muscular activation"
      ],
      "prosthetic_related": [
        "External rotation of the knee, shank, or foot",
        "Socket too tight"
      ]
    },
    "recommended_actions": [
      "Internally rotate the prosthetic knee. Unscrew the component under the socket and rotate the knee.",
      "Internally rotate the prosthetic knee by adjusting the component inferior to the socket.",
      "Rotate the prosthetic knee internally relative to the socket and foot",
      "Internally rotate the prosthetic knee by adjusting the component alignment under the socket.",
      "Internally rotate the prosthetic knee by adjusting the componentry inferior to the socket.",
      "Internally rotate the prosthetic knee by adjusting the rotational component located inferior to the socket."
    ]
  },
  {
    "deviation": "Lateral Whip",
    "general_category": {
      "code": 1,
      "name": "Rotational Deviations"
    },
    "annotation_aliases": [
      "Lateral whip",
      "Internally rotated knee",
      "Internally rotated prosthetic knee"
    ],
    "pattern": "Abrupt lateral movement of the prosthetic heel during the swing phase.",
    "causes": {
      "patient_related": [
        "Socket donned in internal rotation",
        "Gait habit (internal rotation of the hip at toe-off)"
      ],
      "prosthetic_related": [
        "Internal rotation of the knee, shank, or foot"
      ]
    },
    "recommended_actions": [
      "Externally rotate the prosthetic knee. Unscrew the component under the socket and rotate the knee.",
      "Externally rotate the prosthetic knee component.",
      "Externally rotate the prosthetic knee by loosening the proximal component adapter and rotating the knee unit.",
      "Externally rotate the prosthetic knee by adjusting the component under the socket.",
      "Externally rotate the prosthetic knee by adjusting the component inferior to the socket."
    ]
  },
  {
    "deviation": "Excessive Terminal Impact",
    "general_category": {
      "code": 3,
      "name": "Step Length and Timing Issues"
    },
    "annotation_aliases": [
      "Terminal impact",
      "Knee slam",
      "Terminal knee impact",
      "Hard terminal impact after swing extension"
    ],
    "pattern": "The prosthetic knee vigorously extends in terminal swing, causing the foot to visibly or audibly 'slam' into full extension.",
    "causes": {
      "patient_related": [
        "Vigorous hip flexion followed by strong hip extension",
        "Compensatory mechanism because the patient wants to feel the knee fully extend before loading weight"
      ],
      "prosthetic_related": [
        "Insufficient swing control/friction mechanism of the prosthetic knee",
        "Excessive extension assist mechanism"
      ]
    },
    "recommended_actions": [
      "Reduce knee extension assist",
      "Increase knee extension resistance",
      "Reduce swing flexion resistance or decrease extension assist tension",
      "Reduce swing flexion resistance",
      "Reduce knee flexion resistance",
      "Reduce the knee extension assist or increase the knee extension resistance."
    ]
  },
  {
    "deviation": "Increased Knee Flexion (Excessive Heel Rise)",
    "general_category": {
      "code": 4,
      "name": "Knee Instability and Malalignment"
    },
    "annotation_aliases": [
      "Excessive knee flexion",
      "Excessive heel rise"
    ],
    "pattern": "The heel of the prosthesis rises much higher than the sound foot during pre-swing/toe-off.",
    "causes": {
      "patient_related": [
        "Excessive use of strong hip flexors to initiate swing phase, overpowering the knee unit",
        "Hip flexion contracture"
      ],
      "prosthetic_related": [
        "Insufficient stance control mechanism or knee friction",
        "Improper knee selection"
      ]
    }
  },
  {
    "deviation": "Inadequate Knee Flexion in Pre-Swing",
    "general_category": {
      "code": 4,
      "name": "Knee Instability and Malalignment"
    },
    "annotation_aliases": [
      "Insufficient knee flexion",
      "Inadequate knee flexion",
      "Stiff knee in pre-swing",
      "Insufficient swing phase flexion",
      "Insufficient swing flexion",
      "Persistent insufficient swing phase knee flexion."
    ],
    "pattern": "The prosthetic knee does not properly transition to swing phase, remaining too stiff at toe-off.",
    "causes": {
      "patient_related": [
        "Lack of confidence or training in the gait pattern needed to signal the prosthetic knee to switch from stance to swing",
        "Insufficient loading of the toe at the end of stance phase"
      ],
      "prosthetic_related": [
        "Excessive stability of the prosthetic knee"
      ]
    },
    "recommended_actions": [
      "Increase socket flexion, increase plantarflexion of the prosthetic foot, and/or reduce the flexion resistance in the prosthetic knee joint.",
      "Reduce knee flexion resistance by loosening the spring compression in the extension assist mechanism.",
      "Flex the socket and translate it anteriorly",
      "Lower the swing activation threshold of the prosthetic knee, ideally after addressing insufficient prosthesis length and socket flexion.",
      "reduce stance flexion resistance",
      "Reduce the friction resistance of the knee joint."
    ]
  },
  {
    "deviation": "Toe-In Asymmetry / Internally Rotated Foot",
    "general_category": {
      "code": 1,
      "name": "Rotational Deviations"
    },
    "annotation_aliases": [
      "Toe-in asymmetry",
      "Excessive toe-in",
      "Internally rotated foot",
      "Internal foot rotation",
      "Internally rotated prosthetic foot",
      "Internal rotation of the foot",
      "Excessive internal rotation of the prosthetic foot",
      "Excessive toe-in of the prosthetic foot",
      "Excessive toe-in alignment.",
      "Excessive toe-in.",
      "Too much toe-in",
      "Slight toe-in",
      "Too much toe-in coupled with slight inversion or supination"
    ],
    "pattern": "The prosthetic foot progresses with excessive toe-in or internal rotation relative to the sound side during stance or swing.",
    "causes": {
      "patient_related": [
        "Habitual internal hip rotation",
        "Prosthesis donned in internal rotation"
      ],
      "prosthetic_related": [
        "Too much toe-in alignment built into the foot, shank, knee, or socket",
        "Internal rotation of prosthetic components"
      ]
    },
    "recommended_actions": [
      "Externally rotate the prosthetic foot relative to the knee.",
      "Rotate the foot externally relative to the knee.",
      "Adjust the prosthesis to achieve an appropriate degree of toe-out.",
      "Externally rotate the prosthetic foot by loosening the tube clamp and adjusting its rotational position.",
      "Externally rotate the foot to match the contralateral limb's toe-out angle by loosening the tube clamp.",
      "Loosen the tube clamp and rotate the prosthetic foot externally to match the toe-out angle of the contralateral limb."
    ]
  },
  {
    "deviation": "Toe-Out Asymmetry / Externally Rotated Foot",
    "general_category": {
      "code": 1,
      "name": "Rotational Deviations"
    },
    "annotation_aliases": [
      "Toe-out asymmetry",
      "Excessive toe-out",
      "Externally rotated foot",
      "External foot rotation",
      "asymmetric toe-out",
      "Asymmetrical toe-out",
      "Toe-out",
      "Excessive toe-out."
    ],
    "pattern": "The prosthetic foot progresses with excessive toe-out or external rotation relative to the sound side during stance or swing.",
    "causes": {
      "patient_related": [
        "Habitual external hip rotation",
        "Prosthesis donned in external rotation"
      ],
      "prosthetic_related": [
        "Too much toe-out alignment built into the foot, shank, knee, or socket",
        "External rotation of prosthetic components"
      ]
    },
    "recommended_actions": [
      "Rotate prosthetic foot outward by 10-15 degrees",
      "rotate prosthetic foot outward by 10 deg",
      "Rotate the socket externally above the knee",
      "rotate foot outward underneath knee",
      "Externally rotate the prosthetic foot",
      "Correct the alignment by loosening the tube clamp and rotating the foot internally to match the toe-out angle of the sound side."
    ]
  },
  {
    "deviation": "Too Narrow Base of Support",
    "general_category": {
      "code": 2,
      "name": "Step and Base of Support Deviations"
    },
    "annotation_aliases": [
      "Too narrow base of support",
      "Base of support is too narrow",
      "Too narrow step width",
      "Excessively narrow base of support",
      "The base of support is too narrow.",
      "Excessively narrow step width",
      "Insufficient base of support width",
      "Insufficient base of support",
      "Narrow base of support",
      "Excessively narrow base of support."
    ],
    "pattern": "The prosthetic and sound feet are placed too close to the walking midline, narrowing the base of support and risking contact between limbs.",
    "causes": {
      "patient_related": [
        "Habitual narrow walking base",
        "Balance strategy that places the prosthesis too close to midline"
      ],
      "prosthetic_related": [
        "Excessive inset or adduction alignment",
        "Socket or knee alignment that drives the prosthesis toward midline"
      ]
    },
    "recommended_actions": [
      "Increase knee valgus",
      "Adduct the socket by loosening the medial set screw and tightening the lateral set screw inferior to the socket.",
      "Increase knee valgus or shift the socket medially.",
      "Shift socket medially",
      "Adduct the prosthetic socket by loosening the medial set screw and tightening the lateral set screw. An alternative action is to perform a linear outset adjustment of the prosthetic foot.",
      "Adduct the socket. This is accomplished by loosening the medial set screw and tightening the lateral set screw inferior to the socket."
    ]
  },
  {
    "deviation": "Leaning Pylon / Pylon Lean",
    "general_category": {
      "code": 2,
      "name": "Step and Base of Support Deviations"
    },
    "annotation_aliases": [
      "Leaning pylon",
      "Pylon lean",
      "Pylon alignment lean",
      "Leg axis misalignment, indicated by a leaning pylon.",
      "Leaning pylon persists."
    ],
    "pattern": "The prosthetic pylon visibly leans in the frontal plane during stance or progression, changing base width and cosmetic alignment.",
    "causes": {
      "patient_related": [
        "Balance compensation or habitual loading pattern"
      ],
      "prosthetic_related": [
        "Socket/pylon translation or angular alignment error",
        "Foot inset/outset mismatch"
      ]
    },
    "recommended_actions": [
      "Reduce knee valgus and/or translate the socket laterally.",
      "Further reduce knee valgus."
    ]
  },
  {
    "deviation": "Asymmetric Step Length",
    "general_category": {
      "code": 3,
      "name": "Step Length and Timing Issues"
    },
    "annotation_aliases": [
      "Uneven step length",
      "Asymmetric step length",
      "Sound limb step shorter than prosthetic limb step",
      "Prosthetic step length is excessive",
      "The prosthetic foot is positioned too far anteriorly, leading to unequal step lengths."
    ],
    "pattern": "The prosthetic and sound side step lengths are visibly asymmetric outside of normal variation.",
    "causes": {
      "patient_related": [
        "Insufficient hip extension reserve",
        "Pain, weakness, or confidence limitations"
      ],
      "prosthetic_related": [
        "Socket flexion or foot alignment limits smooth progression over the prosthesis"
      ]
    },
    "recommended_actions": [
      "Increase socket flexion",
      "Translate the socket anteriorly",
      "Flex and shift the socket",
      "Flex the socket",
      "Decrease plantarflexion",
      "Translate the foot posteriorly via a linear adjustment."
    ]
  },
  {
    "deviation": "Early Foot Flat",
    "general_category": {
      "code": 3,
      "name": "Step Length and Timing Issues"
    },
    "annotation_aliases": [
      "Early foot flat",
      "Rapid foot flat"
    ],
    "pattern": "The prosthetic foot reaches foot-flat too early after initial contact.",
    "causes": {
      "patient_related": [
        "Cautious loading pattern"
      ],
      "prosthetic_related": [
        "Heel bumper or hydraulic plantarflexion resistance too soft",
        "Excessive plantarflexion alignment"
      ]
    },
    "recommended_actions": [
      "Reduce plantarflexion"
    ]
  },
  {
    "deviation": "Knee Varus Asymmetry",
    "general_category": {
      "code": 4,
      "name": "Knee Instability and Malalignment"
    },
    "annotation_aliases": [
      "Knee varus",
      "Varus deformity",
      "Knee varus asymmetry",
      "Reduce knee varus"
    ],
    "pattern": "The prosthetic knee or socket-pylon alignment presents excessive varus or bow-legged frontal-plane asymmetry.",
    "causes": {
      "patient_related": [
        "Habitual narrow base or frontal-plane balance strategy"
      ],
      "prosthetic_related": [
        "Excessive varus alignment of knee/socket/pylon",
        "Socket/pylon shift used incorrectly to alter step width"
      ]
    },
    "recommended_actions": [
      "Reduce knee varus",
      "reduce varus angle",
      "Reduce the varus moment at the knee by shifting the socket laterally.",
      "Reduce knee varus by shifting the socket laterally."
    ]
  },
  {
    "deviation": "Externally Rotated Knee",
    "general_category": {
      "code": 4,
      "name": "Knee Instability and Malalignment"
    },
    "annotation_aliases": [
      "Externally rotated knee",
      "Knee externally rotated",
      "Rotate knee inward",
      "Excessive external rotation of the prosthetic knee.",
      "Prosthetic knee is excessively externally rotated."
    ],
    "pattern": "The prosthetic knee axis is externally rotated relative to the foot/ankle progression axis.",
    "causes": {
      "patient_related": [
        "Prosthesis donned in external rotation",
        "Habitual external rotation at toe-off"
      ],
      "prosthetic_related": [
        "External rotation of the prosthetic knee, shank, socket, or foot"
      ]
    },
    "recommended_actions": [
      "Internally rotate the prosthetic knee component."
    ]
  },
  {
    "deviation": "Prosthesis Too Short",
    "general_category": {
      "code": 5,
      "name": "Prosthetic Length Issues"
    },
    "annotation_aliases": [
      "Prosthesis is too short",
      "Prosthesis too short",
      "Leg length discrepancy",
      "Stepping into a hole",
      "Prosthesis is too short.",
      "Prosthesis is excessively short.",
      "Prosthetic side is too short.",
      "Prosthetic side is too short, causing a leg length discrepancy.",
      "Prosthesis is too short, creating a leg length discrepancy.",
      "Prosthesis is too short, resulting in a leg length discrepancy.",
      "Prosthesis is too short relative to the contralateral limb.",
      "Prosthetic limb is too short.",
      "Insufficient prosthetic length"
    ],
    "pattern": "The prosthetic limb is physically or functionally short, often producing pelvic drop, lateral trunk compensation, or a feeling of stepping into a hole.",
    "causes": {
      "patient_related": [
        "Residual limb volume loss or incomplete seating changes effective length"
      ],
      "prosthetic_related": [
        "Prosthetic height set too low",
        "Socket fit or suspension reduces effective limb length"
      ]
    },
    "recommended_actions": [
      "Elongate the prosthesis to match the contralateral limb length and ensure a level pelvis.",
      "Lengthen the prosthesis to match the contralateral limb length and ensure a level pelvis.",
      "Extend shank pylon by 5 mm",
      "Lengthen the prosthetic shank by 5-10 mm",
      "Increase the overall length of the prosthesis to match the contralateral limb, thereby leveling the pelvis.",
      "Elongate the prosthesis to match the length of the contralateral limb, ensuring the pelvis is level."
    ]
  },
  {
    "deviation": "Prosthesis Too Long",
    "general_category": {
      "code": 5,
      "name": "Prosthetic Length Issues"
    },
    "annotation_aliases": [
      "Prosthesis is excessively long",
      "Prosthesis too long",
      "Excessively long prosthesis",
      "Shorten the prosthesis",
      "Prosthesis is excessively long.",
      "Excessive prosthetic length.",
      "Excessive prosthetic length causing slight circumduction."
    ],
    "pattern": "The prosthetic limb is physically or functionally long, forcing clearance compensations such as vaulting, circumduction, hip hike, or toe drag.",
    "causes": {
      "patient_related": [
        "Incomplete seating or suspension issues that make the prosthesis functionally long"
      ],
      "prosthetic_related": [
        "Prosthetic height set too high",
        "Excessive plantarflexion or stiff knee behavior increasing functional length"
      ]
    },
    "recommended_actions": [
      "Shorten the prosthesis to match the contralateral limb length and ensure a level pelvis.",
      "Shorten the shank segment by 5-10 mm.",
      "Shorten the prosthesis to match the height of the contralateral limb, thereby leveling the pelvis.",
      "Reduce the prosthetic length to match the contralateral limb and achieve a level pelvis.",
      "Reduce the length of the prosthesis to match the contralateral limb and achieve a level pelvis.",
      "Reduce the prosthesis length to match the contralateral limb and ensure a level pelvis."
    ]
  },
  {
    "deviation": "Incomplete Roll-Over",
    "general_category": {
      "code": 6,
      "name": "Roll-Over and Clearance Issues"
    },
    "annotation_aliases": [
      "Incomplete roll-over",
      "Incomplete rollover",
      "Insufficient roll-over",
      "Lack of confidence loading the prosthesis in late stance",
      "Disrupted swing initiation"
    ],
    "pattern": "Weight transfer over the prosthetic foot is incomplete or hesitant in late stance, disrupting smooth progression to pre-swing.",
    "causes": {
      "patient_related": [
        "Reduced confidence loading the prosthesis late in stance",
        "Habitual avoidance of toe loading"
      ],
      "prosthetic_related": [
        "Socket flexion/translation or foot alignment limits smooth roll-over",
        "Toe lever too long or too stiff"
      ]
    },
    "recommended_actions": [
      "Flex and shift the socket anteriorly.",
      "Flex and shift the socket"
    ]
  },
  {
    "deviation": "Socket Too Wide in ML",
    "general_category": {
      "code": 7,
      "name": "Socket Fit and Stability"
    },
    "annotation_aliases": [
      "Socket too wide in ML",
      "Socket too wide mediolaterally",
      "Poor medial-lateral socket fit",
      "Socket mediolateral dimension is excessive",
      "Alignment is acceptable"
    ],
    "pattern": "The socket appears too wide in the medial-lateral dimension, allowing excess residual-limb or socket motion.",
    "causes": {
      "patient_related": [
        "Residual limb volume loss"
      ],
      "prosthetic_related": [
        "Socket fit too loose in the medial-lateral dimension"
      ]
    },
    "recommended_actions": [
      "Rectify the socket fit"
    ]
  },
  {
    "deviation": "Lateral Instability",
    "general_category": {
      "code": 7,
      "name": "Socket Fit and Stability"
    },
    "annotation_aliases": [
      "Lateral instability",
      "Loss of lateral containment",
      "Socket instability",
      "Lateral knee instability"
    ],
    "pattern": "The prosthetic socket or limb shows unstable lateral motion during stance, often reducing confidence and alignment control.",
    "causes": {
      "patient_related": [
        "Weak hip abductors or poor balance"
      ],
      "prosthetic_related": [
        "Insufficient lateral socket containment",
        "Socket too loose",
        "Incorrect pylon/foot translation"
      ]
    },
    "recommended_actions": [
      "Inset the socket (shift medially) for a transfemoral prosthesis",
      "Correct the varus alignment as previously described",
      "Correct varus alignment as previously described"
    ]
  },
  {
    "deviation": "Incongruity of Knee and Ankle Axes",
    "general_category": {
      "code": 8,
      "name": "Foot and Ankle Deviations"
    },
    "annotation_aliases": [
      "Incongruity of knee and ankle axes",
      "Knee and ankle axes incongruent",
      "Rotate knee inward and foot outward"
    ],
    "pattern": "The prosthetic knee and ankle/foot progression axes are visibly misaligned or rotationally incongruent.",
    "causes": {
      "patient_related": [
        "Habitual gait pattern or donning rotation"
      ],
      "prosthetic_related": [
        "Knee, shank, socket, or foot rotation set inconsistently"
      ]
    },
    "recommended_actions": [
      "Rotate knee inward and foot outward"
    ]
  },
  {
    "deviation": "Excessive Plantar Flexion",
    "general_category": {
      "code": 8,
      "name": "Foot and Ankle Deviations"
    },
    "annotation_aliases": [
      "Excessive plantar flexion",
      "Excessive plantarflexion"
    ],
    "pattern": "The prosthetic foot or ankle is excessively plantarflexed during stance or initial contact.",
    "causes": {
      "patient_related": [
        "Cautious loading pattern"
      ],
      "prosthetic_related": [
        "Foot set in excessive plantarflexion",
        "Plantarflexion bumper or hydraulic resistance mismatch"
      ]
    },
    "recommended_actions": [
      "Reduce plantarflexion"
    ]
  },
  {
    "deviation": "Excessive Dorsiflexion",
    "general_category": {
      "code": 8,
      "name": "Foot and Ankle Deviations"
    },
    "annotation_aliases": [
      "Excessive dorsiflexion",
      "Foot set too dorsiflexed",
      "Excessive dorsiflexion of the prosthetic foot, resulting in the patient's inability to load the toe and a consequent loss of the push-off phase.",
      "Excessive dorsiflexion of the prosthetic foot, preventing the patient from loading the toe and resulting in a loss of the push-off phase."
    ],
    "pattern": "The prosthetic foot or ankle is excessively dorsiflexed, altering heel-off and roll-over.",
    "causes": {
      "patient_related": [
        "Compensatory loading pattern"
      ],
      "prosthetic_related": [
        "Foot set in excessive dorsiflexion",
        "Toe lever too short or keel too soft"
      ]
    },
    "recommended_actions": [
      "Induce plantarflexion of the prosthetic foot. This is achieved by loosening the anterior set screw and tightening the posterior set screw located superior to the ankle component.",
      "Increase plantarflexion of the prosthetic foot by loosening the anterior set screw and tightening the posterior set screw of the ankle component."
    ]
  },
  {
    "deviation": "Pronated or Everted Foot",
    "general_category": {
      "code": 8,
      "name": "Foot and Ankle Deviations"
    },
    "annotation_aliases": [
      "Pronated foot",
      "Everted foot",
      "Excessive foot eversion",
      "Initial contact on the medial border",
      "Pronated or everted foot",
      "Excessive foot eversion, causing the prosthesis to land on its medial border.",
      "Excessive foot eversion, causing the prosthesis to land on the medial border.",
      "Excessive foot eversion, causing the prosthesis to make initial contact on its medial border.",
      "Excessive foot eversion, causing the prosthesis to make initial contact on the medial border.",
      "Excessive foot eversion, causing initial contact on the medial border of the prosthetic foot.",
      "Excessive foot eversion, causing the prosthesis to contact the ground on its medial border."
    ],
    "pattern": "The prosthetic foot is excessively pronated or everted, sometimes contacting on the medial border at initial contact.",
    "causes": {
      "patient_related": [
        "Habitual foot-placement compensation"
      ],
      "prosthetic_related": [
        "Foot inversion/eversion alignment error",
        "Ankle or foot component set in excessive eversion"
      ]
    },
    "recommended_actions": [
      "Invert the prosthetic foot by loosening the lateral set screw and tightening the medial set screw superior to the ankle component.",
      "Supinate or invert the foot.",
      "Loosen the lateral set screw and tighten the medial set screw superior to the foot.",
      "Invert the foot. Loosen the lateral set screw and tighten the medial set screw superior to the ankle component.",
      "Perform a foot inversion alignment adjustment. Loosen the lateral set screw and tighten the medial set screw located superior to the ankle component.",
      "Invert the foot by loosening the lateral set screw and tightening the medial set screw superior to the ankle component."
    ]
  },
  {
    "deviation": "Normal Gait",
    "general_category": {
      "code": 9,
      "name": "Normal Gait"
    },
    "annotation_aliases": [
      "Normal gait",
      "No action needed",
      "Optimized gait",
      "Acceptable gait",
      "Natural",
      "Natural gait",
      "Natural gait pattern",
      "Patient's gait is consistent with their established pattern."
    ],
    "pattern": "No clinically meaningful prosthetic gait deviation is supported by the available visual and metric evidence.",
    "causes": {
      "patient_related": [],
      "prosthetic_related": []
    },
    "recommended_actions": [
      "No action needed",
      "Knee fine-tuning to be determined outside of the parallel bars",
      "Consider fine-tuning the prosthesis after an extended accommodation period.",
      "Observe further",
      "Further observation is recommended",
      "No intervention required"
    ]
  },
  {
    "deviation": "Input Not Suitable for Analysis",
    "general_category": {
      "code": 0,
      "name": "Input Not Suitable for Analysis"
    },
    "annotation_aliases": [
      "Input not suitable for analysis",
      "No action can be determined",
      "Provided text not suitable for gait analysis"
    ],
    "pattern": "The input does not contain enough valid gait-analysis information to assign a prosthetic gait deviation.",
    "causes": {
      "patient_related": [],
      "prosthetic_related": []
    },
    "recommended_actions": [
      "No action can be determined"
    ]
  },
  {
    "deviation": "Inconclusive Frontal Plane Alignment Assessment",
    "general_category": {
      "code": 99,
      "name": "Inconclusive or Unscorable Gait Assessment"
    },
    "annotation_aliases": [
      "Assessment of frontal plane alignment (varus/valgus) was inconclusive",
      "Recommend re-evaluation with a camera providing a direct frontal plane view"
    ],
    "pattern": "The available view is insufficient to assess the intended frontal-plane alignment change, such as varus or valgus.",
    "causes": {
      "patient_related": [],
      "prosthetic_related": []
    },
    "recommended_actions": [
      "Recommend re-evaluation with a camera providing a direct frontal plane view"
    ]
  },
  {
    "deviation": "Excessive Socket Adduction / Valgus Gait",
    "general_category": {
      "code": 2,
      "name": "Step and Base of Support Deviations"
    },
    "annotation_aliases": [
      "Excessive socket adduction (valgus gait)",
      "Excessive socket adduction (valgus gait), causing initial foot contact on the medial border with the pylon exhibiting a medial lean.",
      "Excessive socket adduction, resulting in a valgus gait with medial foot contact and a medially inclined pylon.",
      "Excessive socket adduction resulting in a valgus gait pattern, causing the foot to land on its medial border and the pylon to lean medially.",
      "Excessive socket adduction, resulting in a valgus gait pattern with medial pylon leaning.",
      "Excessive socket adduction (valgus gait), resulting in a wide base of support.",
      "Excessive socket adduction (valgus gait) resulting in a wide base of support.",
      "Excessive socket adduction, valgus gait",
      "Excessive socket adduction",
      "Excessive adduction of the socket.",
      "Medially inclined pylon",
      "Pylon exhibiting a medial lean"
    ],
    "pattern": "The socket is excessively adducted, producing a valgus gait pattern, medial pylon lean, medial foot contact, or abnormal base width.",
    "causes": {
      "patient_related": [
        "Compensatory frontal-plane loading pattern"
      ],
      "prosthetic_related": [
        "Socket adduction alignment is excessive",
        "Socket/pylon/foot alignment produces a valgus gait pattern"
      ]
    },
    "recommended_actions": [
      "Abduct the socket by loosening the lateral set screw and tightening the medial set screw.",
      "Abduct the socket. This is achieved by loosening the lateral set screw and tightening the medial set screw on the pyramid adapter inferior to the socket.",
      "Abduct the socket by loosening the lateral set screw and tightening the medial set screw at the socket adapter.",
      "Induce foot eversion by loosening the medial set screw and tightening the lateral set screw at the ankle adapter.",
      "Induce foot dorsiflexion by loosening the posterior set screw and tightening the anterior set screw at the ankle adapter.",
      "Introduce a slight amount of socket abduction by loosening the lateral set screw and tightening the medial set screw of the pyramid adapter."
    ]
  },
  {
    "deviation": "Foot Inversion / Supinated Foot",
    "general_category": {
      "code": 8,
      "name": "Foot and Ankle Deviations"
    },
    "annotation_aliases": [
      "Foot inversion",
      "Inverted foot",
      "Supinated foot",
      "Associated foot inversion and plantarflexion.",
      "Slightly inverted foot causing the prosthesis to land on its lateral border.",
      "Slight foot inversion causing initial contact on the lateral border of the prosthetic foot.",
      "Slight foot inversion at initial contact, causing the prosthesis to land on its lateral border.",
      "Too much toe-in coupled with slight inversion or supination",
      "Initial contact on the lateral border of the prosthetic foot",
      "Excessive foot inversion, causing initial contact on the lateral border of the prosthetic foot."
    ],
    "pattern": "The prosthetic foot is excessively inverted or supinated, often contacting on the lateral border at initial contact.",
    "causes": {
      "patient_related": [
        "Habitual inverted foot placement or toe-in compensation"
      ],
      "prosthetic_related": [
        "Foot/ankle component set in excessive inversion or supination",
        "Socket/foot alignment drives lateral-border initial contact"
      ]
    },
    "recommended_actions": [
      "Toe-out and foot eversion or pronation",
      "Abduct the socket by loosening the lateral set screw and tightening the medial set screw at the socket adapter.",
      "Induce foot eversion by loosening the medial set screw and tightening the lateral set screw at the ankle adapter.",
      "Induce foot dorsiflexion by loosening the posterior set screw and tightening the anterior set screw at the ankle adapter.",
      "Adjust the foot into eversion. Loosen the medial set screw and tighten the lateral set screw located superior to the ankle component.",
      "Correct for foot inversion by loosening the medial set screw and tightening the lateral set screw above the ankle."
    ]
  },
  {
    "deviation": "Asymmetric Knee Flexion Angles",
    "general_category": {
      "code": 8,
      "name": "Foot and Ankle Deviations"
    },
    "annotation_aliases": [
      "Asymmetric knee flexion angles"
    ],
    "pattern": "The prosthetic and contralateral knees show asymmetric flexion angles that affect swing or stance mechanics.",
    "causes": {
      "patient_related": [
        "Activity-level or confidence mismatch affecting knee use"
      ],
      "prosthetic_related": [
        "Swing flexion/extension resistance settings are too high or otherwise mismatched"
      ]
    },
    "recommended_actions": [
      "Lower swing flexion resistance"
    ]
  },
  {
    "deviation": "Abducted and Externally Rotated Prosthetic Knee",
    "general_category": {
      "code": 4,
      "name": "Knee Instability and Malalignment"
    },
    "annotation_aliases": [
      "Prosthetic knee is abducted and externally rotated.",
      "Abducted and externally rotated prosthetic knee",
      "Prosthetic knee is abducted and externally rotated"
    ],
    "pattern": "The prosthetic knee is malaligned in abduction and external rotation relative to the limb progression line.",
    "causes": {
      "patient_related": [
        "Habitual rotated or abducted limb progression"
      ],
      "prosthetic_related": [
        "Knee component rotation and linear alignment are incorrect"
      ]
    },
    "recommended_actions": [
      "Internally rotate the knee component to correct for the external rotation.",
      "Shift the knee joint medially via a linear adjustment to address the abduction, which is an embedded design feature.",
      "Internally rotate the knee to address the external rotation. To correct the abduction, perform a medial linear shift of the knee joint. Addressing the knee joint alignment is a prerequisite for correcting the overall abducted gait."
    ]
  }
]
\end{lstlisting}

\section{Gait Category Mapping for GAVD}
\label{sec:gavd_mapping}
\begin{itemize}
    \item \textbf{Normal:} Healthy (Normal) Gait
    \item \textbf{Antalgic:} Antalgic Gait
    \item \textbf{Stroke:} Hemiparetic (Hemiplegic) Gait
    \item \textbf{Parkinson's:} Parkinsonian (Festinating) Gait
    \item \textbf{Myopathic:} Waddling (Myopathic) Gait, Trendelenburg Gait
    \item \textbf{Cerebral Palsy:} Spastic Diplegic (Scissors) Gait, Crouch Gait, Spastic Paraparetic Gait, Hemiparetic (Hemiplegic) Gait
\end{itemize}

\section{DrGait Prompt Templates}
\label{sec:prompts}
\begin{lstlisting}[style=plaintext, caption={System Prompt for GAVD}]
You are the reasoning module in a Triage-Verification-Synthesis (TVS) gait-analysis pipeline.

You can be invoked for different stages:
- TRIAGE: first decide normal-vs-abnormal gait, then propose the initial hypothesis path and measurable verification objectives.
- VERIFICATION: use deterministic tools to confirm/refute objectives and hypotheses.
- SYNTHESIS: integrate verified evidence into a final, transparent conclusion.

Global policy:
- Never invent numerical values, thresholds, or measurements.
- Ground quantitative claims in provided metrics and/or tool outputs.
- Do NOT use a single evidence to rule out a hypothesis.
- Joint angles are measured in degrees where ~180$^\circ$ is fully extended (straight).

Tool policy:
- When tools are available, call tools for measurable biomechanical claims.
- Use only declared tools and keep each call purposeful.

Output policy:
- Follow the stage-specific output schema requested in the user prompt.
- During TRIAGE, do not jump directly to differential diagnosis before the normality gate.
- Return only valid JSON when a schema is requested.
- Do not output extra prose outside the requested format.
\end{lstlisting}

\begin{lstlisting}[style=plaintext, caption={Triage User Prompt for GAVD}]
Stage: TRIAGE

Analyze the attached walking video and perform TVS triage.

Quantitative metrics:
{metrics_table}

Task:
1) First perform a normality gate: decide whether the gait is `normal`, `abnormal`, or `uncertain`.
2) Then propose the initial candidate hypotheses `H_path` according to that gate.
3) Propose explicit biomechanical verification objectives `C_bio` for downstream tool execution, focused on visual or numerical uncertainty.

Constraints:
- Candidate labels must come from the injected gait knowledge dictionary.
- Use category names exactly as listed in the dictionary.
- Use BOTH visual signs and provided metrics.
- Rhythm caution: High variation in step_length may indicate a lack of rhythm, but it may also come from problematic metric measurements. Do not use it alone to confirm or rule out any hypothesis.
- The `normal` hypothesis category is exactly `Healthy (Normal) Gait`.
- If the normality gate is `normal`, `H_path` MUST contain exactly one entry: `Healthy (Normal) Gait`. Do not include abnormal alternatives in `H_path`.
- If the normality gate is `abnormal`, `H_path` should contain only abnormal candidate categories; do not include `Healthy (Normal) Gait`.
- If the normality gate is `uncertain`, choose the safest initial hypothesis path based on current evidence, and use `C_bio` to verify the uncertainty before escalation.
- Even when `H_path` contains only `Healthy (Normal) Gait`, `C_bio` must still include tool-check objectives to inspect visual or numerical uncertainty, such as subtle asymmetry, step width/variability, small steps, stiff knee, circumduction, sway, or posture.
- MEASURING SWAY: To verify Waddling or Ataxic gaits, you MUST use `get_postural_sway_amplitude` to measure the dynamic range. Do not use single-frame angles.
- Do not invent numerical values, thresholds, or measurements.

Return format:
- Return valid JSON only.
- Use this exact top-level schema:
{
  "normality_gate": {
    "classification": "normal|abnormal|uncertain",
    "evidence": [
      "visual or metric-grounded evidence used for the normal-vs-abnormal gate"
    ],
    "uncertainty": [
      "visual or numerical ambiguity that should be checked downstream"
    ]
  },
  "H_path":[
    {
      "id": 1,
      "category": "exact clinical category string from injected knowledge dictionary",
      "confidence": 0.0,
      "evidence":[
        "metric-grounded or visual sign",
        "metric-grounded or visual sign"
      ],
      "uncertainty":[
        "missing cue, conflicting metric, or ambiguity"
      ]
    }
  ],
  "C_bio":[
    {
      "id": 1,
      "objective": "biomechanical claim to verify",
      "target_hypotheses": ["category_a", "category_b"],
      "priority": "high",
      "required_observables":[
        "what to measure from tools (angles, asymmetry, etc.)"
      ],
      "decision_rule": "what outcome would support vs refute target hypotheses"
    }
  ],
  "ruled_out_distractors":[
    "optional: dictionary categories considered but excluded after triage"
  ],
  "triage_notes": "1-3 sentences summarizing differential logic and main uncertainty"
}
- `H_path` must contain 1-5 entries with confidence in [0,1] and non-increasing by rank.
- If `normality_gate.classification` is `normal`, `H_path` must contain exactly one entry and its category must be `Healthy (Normal) Gait`.
- `C_bio` must contain 1-10 objectives; `priority` must be one of: high, medium, low.

{extra_context_block}

\end{lstlisting}

\begin{lstlisting}[style=plaintext, caption={Verification User Prompt for GAVD}]
Stage: VERIFICATION

Run TVS verification for this gait case.

Instructions:
- Use declared function tools to verify the triage objectives and key differential points.
- Prioritize high-priority objectives first.
- This stage allows only one tool-calling round.
- Each function call must include why_used, explicitly stating which differential diagnosis it aims to resolve (e.g., 'Measuring step width to differentiate Ataxic vs. Hemiplegic').

{extra_context_block}

\end{lstlisting}

\begin{lstlisting}[style=plaintext, caption={Synthesis User Prompt for GAVD}]
Stage: SYNTHESIS

Use the full TVS conversation (triage hypotheses, verification tool calls and results) to produce the recursive synthesis.

Instructions:
- First summarize the verification objectives in `C_bio` using only the tool results already in context.
- In each `C_bio[*].evidence` item, briefly state which tool produced the evidence and what key result supports, refutes, or leaves the objective inconclusive.
- Then determine `H_path` status from the `C_bio` evidence.
- Explicitly identify contradictions between triage claims and tool-grounded findings.
- Do not invent measurements; cite only values/observations present in prior context.
- If evidence is insufficient or conflicting, state this clearly.
- The response should be extend from last triage output. Do not invent new triage hypotheses or verification objectives. Focus on synthesizing the existing information.
- `H_path` may contain 1-5 entries with confidence in [0,1] and non-increasing by rank.
- Write `final_report` as the user-facing conclusion: state the final gait-category conclusion, briefly explain why the decision was made, and include a concise suggested action or next step to address the gait finding. Keep the action general and evidence-grounded, such as clinical follow-up, targeted gait training, strength/balance work, assistive-device review, or additional assessment when uncertainty remains.

Return format:
- Return valid JSON only.
- Use this exact schema:
{
  "C_bio":[
    {
      "id": 1,
      "objective": "biomechanical claim to verify",
      "target_hypotheses":["category_a", "category_b"],
      "status": "supported|refuted|inconclusive",
      "evidence": ["tool name + key result and interpretation", "..."]
    }
  ],
  "H_path":[
    {
      "id": 1,
      "category": "exact clinical category string from injected knowledge dictionary",
      "supporting_evidence":["C_bio-grounded statement", "..."],
      "conflicting_evidence": ["contradiction or empty list item if none"],
      "status": "confirmed|rejected|inconclusive",
      "confidence": 0.0
    }
  ],
  "synthesis_notes": "1-3 sentences summarizing new findings, contradictions, and remaining uncertainty. This will be added to the conversation context for next iteration.",
  "final_report": "User-facing final conclusion, brief rationale for the decision, and concise suggested action or next step."
}

{extra_context_block}

\end{lstlisting}

\section{Tool Declarations}
\label{sec:tools}
\begin{lstlisting}[style=plaintext, caption={GAVD Tools}]
[
    {
        "name": "get_gait_event_frame",
        "description": "Return up to 5 raw video RGB frames at exact gait-event frames from complete gait cycles only. Each entry includes one base64 JPEG in `frame` plus frame metadata.",
        "parameters": {
            "type": "object",
            "properties": {
                "event": {"type": "string", "enum": VALID_EVENT_TOKENS},
                "step_index": {"type": "integer"},
                "why_used": {
                    "type": "string",
                    "description": "One-sentence biomechanical reason for this tool call.",
                },
            },
            "required": ["event", "why_used"],
        },
    },
    {
        "name": "get_joint_angle_at_event",
        "description": "Get joint angle at specified gait event from complete gait cycles only. Cross-side combinations are allowed (e.g., left_knee at right_initial_contact).",
        "parameters": {
            "type": "object",
            "properties": {
                "joint": {
                    "type": "string",
                    "enum": VALID_ANGLE_JOINT_TOKENS,
                    "description": "Sided joint token, e.g. left_knee.",
                },
                "plane": {"type": "string", "enum": ["default", "sagittal", "coronal", "transverse"]},
                "event": {"type": "string", "enum": VALID_EVENT_TOKENS},
                "why_used": {
                    "type": "string",
                    "description": "One-sentence biomechanical reason for this tool call.",
                },
            },
            "required": ["joint", "plane", "event", "why_used"],
        },
    },
    {
        "name": "get_peak_joint_angle",
        "description": "Get overall max/min joint angles in complete-cycle portions of a gait phase. Cross-side combinations are allowed (e.g., left_knee during right_stance).",
        "parameters": {
            "type": "object",
            "properties": {
                "joint": {
                    "type": "string",
                    "enum": VALID_ANGLE_JOINT_TOKENS,
                    "description": "Sided joint token, e.g. left_knee.",
                },
                "plane": {"type": "string", "enum": ["sagittal", "coronal", "transverse"]},
                "phase": {"type": "string", "enum": VALID_PHASE_TOKENS},
                "why_used": {
                    "type": "string",
                    "description": "One-sentence biomechanical reason for this tool call.",
                },
            },
            "required": ["joint", "plane", "phase", "why_used"],
        },
    },
    {
        "name": "get_postural_angle",
        "description": "Get the unsigned body-oriented posture angle of pelvis or trunk at ONE specific frame. For trunk, this is the angle between the trunk segment and body vertical in the requested plane. 0 or 180 degrees means aligned/straight with the reference axis; 90 degrees means perpendicular. WARNING: Do not use this to measure dynamic rhythmic sway (like Waddling). Use get_postural_sway_amplitude for dynamic range.",
        "parameters": {
            "type": "object",
            "properties": {
                "segment": {"type": "string", "enum": ["pelvis", "trunk"]},
                "plane": {"type": "string", "enum": ["sagittal", "coronal"]},
                "frame": {"type": "integer"},
                "why_used": {
                    "type": "string",
                    "description": "One-sentence biomechanical reason for this tool call."
                }
            },
            "required":["segment", "plane", "frame", "why_used"]
        }
    },
    {
        "name": "get_arm_swing_amplitude",
        "description": "Get signed shoulder arm-swing extrema over complete gait cycles for one side in the body-oriented sagittal plane. Returns max_angle_deg as anterior swing (positive) and min_angle_deg as posterior swing (negative).",
        "parameters": {
            "type": "object",
            "properties": {
                "side": {"type": "string", "enum": ["left", "right"]},
                "why_used": {
                    "type": "string",
                    "description": "One-sentence biomechanical reason for this tool call.",
                },
            },
            "required": ["side", "why_used"],
        },
    },
    {
        "name": "get_foot_clearance",
        "description": "Get minimum foot clearance from complete-cycle portions of the requested phase, normalized by body height.",
        "parameters": {
            "type": "object",
            "properties": {
                "phase": {"type": "string", "enum": VALID_PHASE_TOKENS},
                "why_used": {
                    "type": "string",
                    "description": "One-sentence biomechanical reason for this tool call.",
                },
            },
            "required": ["phase", "why_used"],
        },
    },
    {
        "name": "get_lateral_deviation",
        "description": "Get maximum lateral deviation of a sided lower-limb joint from complete-cycle portions of the requested phase, normalized by shoulder width. Cross-side combinations are allowed (e.g., left_foot during right_stance).",
        "parameters": {
            "type": "object",
            "properties": {
                "joint": {"type": "string", "enum": VALID_LATERAL_JOINT_TOKENS},
                "phase": {"type": "string", "enum": VALID_PHASE_TOKENS},
                "why_used": {
                    "type": "string",
                    "description": "One-sentence biomechanical reason for this tool call.",
                },
            },
            "required": ["joint", "phase", "why_used"],
        },
    },
    {
        "name": "get_vertical_com_displacement",
        "description": "Get total vertical COM displacement normalized by body height, restricted to complete gait cycles.",
        "parameters": {
            "type": "object",
            "properties": {
                "why_used": {
                    "type": "string",
                    "description": "One-sentence biomechanical reason for this tool call.",
                },
            },
            "required": ["why_used"],
        },
    },
    {
        "name": "get_postural_sway_amplitude",
        "description": "Get the dynamic range of motion (Max Angle - Min Angle) of the trunk or pelvis over complete gait cycles. Essential for diagnosing Waddling (coronal sway), Ataxic instability, or Trendelenburg compensations.",
        "parameters": {
            "type": "object",
            "properties": {
                "segment": {"type": "string", "enum": ["pelvis", "trunk"]},
                "plane": {"type": "string", "enum": ["sagittal", "coronal"]},
                "why_used": {
                    "type": "string",
                    "description": "One-sentence biomechanical reason for this tool call."
                }
            },
            "required": ["segment", "plane", "why_used"]
        }
    },
    {
        "name": "calculate_symmetry_index",
        "description": "Calculate bilateral symmetry index in percent for step_length, stance_time, or swing_time.",
        "parameters": {
            "type": "object",
            "properties": {
                "metric_name": {"type": "string", "enum": ["step_length", "stance_time", "swing_time"]},
                "why_used": {
                    "type": "string",
                    "description": "One-sentence biomechanical reason for this tool call."
                }
            },
            "required": ["metric_name", "why_used"]
        }
    },
]

\end{lstlisting}

\begin{lstlisting}[style=plaintext, caption={ProGait Tools}]
[
    {
        "name": "get_gait_event_frame",
        "description": "Return up to 5 raw video RGB frames at exact gait-event frames from complete gait cycles only. In segmented 2D contexts, set view to one of front, back, left, or right when the requested evidence is segment-specific. Each entry includes one base64 JPEG in `frame` plus frame metadata.",
        "parameters": {
            "type": "object",
            "properties": {
                "event": {"type": "string", "enum": VALID_EVENT_TOKENS},
                "step_index": {"type": "integer"},
                "view": {"type": "string", "enum": PROGAIT_SEGMENT_VIEW_TOKENS},
                "why_used": {
                    "type": "string",
                    "description": "One-sentence biomechanical reason for this tool call.",
                },
            },
            "required": ["event", "why_used"],
        },
    },
    {
        "name": "get_joint_angle_at_event",
        "description": "Get joint angle at specified gait event from complete gait cycles only. In image-based 2D contexts, this returns an image-plane angle from the selected view rather than calibrated 3D rotation; ankle angles use the average of toe and little-toe landmarks when explicit foot joints are not available. Cross-side combinations are allowed (e.g., left_knee at right_initial_contact).",
        "parameters": {
            "type": "object",
            "properties": {
                "joint": {
                    "type": "string",
                    "enum": VALID_ANGLE_JOINT_TOKENS,
                    "description": "Sided joint token, e.g. left_knee.",
                },
                "plane": {"type": "string", "enum": ["default", "sagittal", "coronal", "transverse"]},
                "event": {"type": "string", "enum": VALID_EVENT_TOKENS},
                "view": {"type": "string", "enum": PROGAIT_SEGMENT_VIEW_TOKENS},
                "why_used": {
                    "type": "string",
                    "description": "One-sentence biomechanical reason for this tool call.",
                },
            },
            "required": ["joint", "plane", "event", "why_used"],
        },
    },
    {
        "name": "get_peak_joint_angle",
        "description": "Get overall max/min joint angles in complete-cycle portions of a gait phase. In image-based 2D contexts, this returns image-plane extrema from the selected view rather than calibrated 3D rotation; ankle angles use the average of toe and little-toe landmarks when explicit foot joints are not available. Cross-side combinations are allowed (e.g., left_knee during right_stance).",
        "parameters": {
            "type": "object",
            "properties": {
                "joint": {
                    "type": "string",
                    "enum": VALID_ANGLE_JOINT_TOKENS,
                    "description": "Sided joint token, e.g. left_knee.",
                },
                "plane": {"type": "string", "enum": ["sagittal", "coronal", "transverse"]},
                "phase": {"type": "string", "enum": VALID_PHASE_TOKENS},
                "view": {"type": "string", "enum": PROGAIT_SEGMENT_VIEW_TOKENS},
                "why_used": {
                    "type": "string",
                    "description": "One-sentence biomechanical reason for this tool call.",
                },
            },
            "required": ["joint", "plane", "phase", "why_used"],
        },
    },
    {
        "name": "get_postural_angle",
        "description": "Get the unsigned body-oriented posture angle of pelvis or trunk at ONE specific frame. For trunk, this is the angle between the trunk segment and body vertical in the requested plane. 0 or 180 degrees means aligned/straight with the reference axis; 90 degrees means perpendicular. WARNING: Do not use this to measure dynamic rhythmic sway (like Waddling). Use get_postural_sway_amplitude for dynamic range.",
        "parameters": {
            "type": "object",
            "properties": {
                "segment": {"type": "string", "enum": ["pelvis", "trunk"]},
                "plane": {"type": "string", "enum": ["sagittal", "coronal"]},
                "frame": {"type": "integer"},
                "view": {"type": "string", "enum": PROGAIT_SEGMENT_VIEW_TOKENS},
                "why_used": {
                    "type": "string",
                    "description": "One-sentence biomechanical reason for this tool call."
                }
            },
            "required":["segment", "plane", "frame", "why_used"]
        }
    },
    {
        "name": "get_arm_swing_amplitude",
        "description": "Get signed shoulder arm-swing extrema over complete gait cycles for one side. In segmented 2D contexts, use the view that best represents sagittal-plane arm motion when available.",
        "parameters": {
            "type": "object",
            "properties": {
                "side": {"type": "string", "enum": ["left", "right"]},
                "view": {"type": "string", "enum": PROGAIT_SEGMENT_VIEW_TOKENS},
                "why_used": {
                    "type": "string",
                    "description": "One-sentence biomechanical reason for this tool call.",
                },
            },
            "required": ["side", "why_used"],
        },
    },
    {
        "name": "get_foot_clearance",
        "description": "Get minimum foot clearance from complete-cycle portions of the requested phase, normalized by body height. In image-based 2D contexts the result is an image-plane proxy.",
        "parameters": {
            "type": "object",
            "properties": {
                "phase": {"type": "string", "enum": VALID_PHASE_TOKENS},
                "view": {"type": "string", "enum": PROGAIT_SEGMENT_VIEW_TOKENS},
                "why_used": {
                    "type": "string",
                    "description": "One-sentence biomechanical reason for this tool call.",
                },
            },
            "required": ["phase", "why_used"],
        },
    },
    {
        "name": "get_lateral_deviation",
        "description": "Get maximum lateral deviation of a sided lower-limb joint from complete-cycle portions of the requested phase, normalized by shoulder width. In image-based 2D contexts the result is an image-plane proxy. Cross-side combinations are allowed.",
        "parameters": {
            "type": "object",
            "properties": {
                "joint": {"type": "string", "enum": VALID_LATERAL_JOINT_TOKENS},
                "phase": {"type": "string", "enum": VALID_PHASE_TOKENS},
                "view": {"type": "string", "enum": PROGAIT_SEGMENT_VIEW_TOKENS},
                "why_used": {
                    "type": "string",
                    "description": "One-sentence biomechanical reason for this tool call.",
                },
            },
            "required": ["joint", "phase", "why_used"],
        },
    },
    {
        "name": "get_foot_progression_angle",
        "description": "Estimate foot progression/orientation from heel-to-toe landmarks during a complete-cycle phase. In image-based 2D contexts this is an image-plane proxy for toe-in/toe-out and should be compared side-to-side or with event frames.",
        "parameters": {
            "type": "object",
            "properties": {
                "side": {"type": "string", "enum": ["left", "right"]},
                "phase": {"type": "string", "enum": ["stance", "swing", "full_cycle"]},
                "view": {"type": "string", "enum": PROGAIT_SEGMENT_VIEW_TOKENS},
                "why_used": {
                    "type": "string",
                    "description": "One-sentence biomechanical reason for this tool call.",
                },
            },
            "required": ["side", "why_used"],
        },
    },
    {
        "name": "get_whip_direction",
        "description": "Estimate whether the heel moves medially or laterally during swing using complete-cycle heel trajectory. Use for medial/lateral whip hypotheses and confirm abruptness with event frames when needed.",
        "parameters": {
            "type": "object",
            "properties": {
                "side": {"type": "string", "enum": ["left", "right"]},
                "view": {"type": "string", "enum": PROGAIT_SEGMENT_VIEW_TOKENS},
                "why_used": {
                    "type": "string",
                    "description": "One-sentence biomechanical reason for this tool call.",
                },
            },
            "required": ["side", "why_used"],
        },
    },
    {
        "name": "get_knee_ankle_axis_alignment",
        "description": "Estimate frontal-plane knee-to-ankle axis alignment during a complete-cycle phase. Use for varus/valgus, pylon lean, and knee/ankle axis incongruity hypotheses.",
        "parameters": {
            "type": "object",
            "properties": {
                "side": {"type": "string", "enum": ["left", "right"]},
                "phase": {"type": "string", "enum": ["stance", "swing", "full_cycle"]},
                "view": {"type": "string", "enum": PROGAIT_SEGMENT_VIEW_TOKENS},
                "why_used": {
                    "type": "string",
                    "description": "One-sentence biomechanical reason for this tool call.",
                },
            },
            "required": ["side", "why_used"],
        },
    },
    {
        "name": "get_vertical_com_displacement",
        "description": "Get total vertical body-center displacement normalized by body height, restricted to complete gait cycles. In image-based 2D contexts the result is an image-plane proxy.",
        "parameters": {
            "type": "object",
            "properties": {
                "view": {"type": "string", "enum": PROGAIT_SEGMENT_VIEW_TOKENS},
                "why_used": {
                    "type": "string",
                    "description": "One-sentence biomechanical reason for this tool call.",
                },
            },
            "required": ["why_used"],
        },
    },
    {
        "name": "get_postural_sway_amplitude",
        "description": "Get the dynamic range of motion (Max Angle - Min Angle) of the trunk or pelvis over complete gait cycles. In segmented 2D contexts, use the view that best represents the requested anatomical plane.",
        "parameters": {
            "type": "object",
            "properties": {
                "segment": {"type": "string", "enum": ["pelvis", "trunk"]},
                "plane": {"type": "string", "enum": ["sagittal", "coronal"]},
                "view": {"type": "string", "enum": PROGAIT_SEGMENT_VIEW_TOKENS},
                "why_used": {
                    "type": "string",
                    "description": "One-sentence biomechanical reason for this tool call."
                }
            },
            "required": ["segment", "plane", "why_used"]
        }
    },
    {
        "name": "calculate_symmetry_index",
        "description": "Calculate bilateral symmetry index in percent for step_length, stance_time, or swing_time. In multiview 2D contexts, left/right segment apparent step-length and timing metrics are averaged when available.",
        "parameters": {
            "type": "object",
            "properties": {
                "metric_name": {"type": "string", "enum": ["step_length", "stance_time", "swing_time"]},
                "why_used": {
                    "type": "string",
                    "description": "One-sentence biomechanical reason for this tool call."
                }
            },
            "required": ["metric_name", "why_used"]
        }
    },
]
\end{lstlisting}



\end{document}